\documentclass[journal]{IEEEtran}
\usepackage{amsmath,amsfonts}
\usepackage{algorithmic}
\usepackage{algorithm}
\usepackage{array}
\usepackage[caption=false,font=normalsize,labelfont=sf,textfont=sf]{subfig}
\usepackage{textcomp}
\usepackage{stfloats}
\usepackage{url}
\usepackage{verbatim}
\usepackage{graphicx}
\usepackage{cite}

\usepackage{setspace}
\usepackage{multirow}
\usepackage{makecell}
\usepackage{booktabs}
\usepackage{color}
\usepackage{url}

\definecolor{rose}{RGB}{255,56,255} 

\begin{document}

\title{How is Visual Attention Influenced by Text \\ Guidance? Database and Model}

\author{Yinan~Sun,
        Xiongkuo~Min$^{*}$,~\IEEEmembership{Member,~IEEE,}
        Huiyu~Duan,
        and~Guangtao~Zhai$^{*}$,~\IEEEmembership{Senior Member,~IEEE}
\thanks{$^{*}$Corresponding authors.}
}

\maketitle

\begin{abstract}
The analysis and prediction of visual attention have long been crucial tasks in the fields of computer vision and image processing. In practical applications, images are generally accompanied by various text descriptions, however, few studies have explored the influence of text descriptions on visual attention, let alone developed visual saliency prediction models considering text guidance. In this paper, we conduct a comprehensive study on text-guided image saliency (TIS) from both subjective and objective perspectives. Specifically, we construct a TIS database named SJTU-TIS, which includes 1200 text-image pairs and the corresponding collected eye-tracking data. Based on the established SJTU-TIS database, we analyze the influence of various text descriptions on visual attention. Then, to facilitate the development of saliency prediction models considering text influence, we construct a benchmark for the established SJTU-TIS database using state-of-the-art saliency models. Finally, considering the effect of text descriptions on visual attention, while most existing saliency models ignore this impact, we further propose a text-guided saliency (TGSal) prediction model, which extracts and integrates both image features and text features to predict the image saliency under various text-description conditions. Our proposed model significantly outperforms the state-of-the-art saliency models on both the SJTU-TIS database and the pure image saliency databases in terms of various evaluation metrics. The SJTU-TIS database and the code of the proposed TGSal model will be released at: {\color{rose}https://github.com/IntMeGroup/TGSal.}
\end{abstract}


\begin{figure}[!t]
\centering
\includegraphics[width=2.8in]{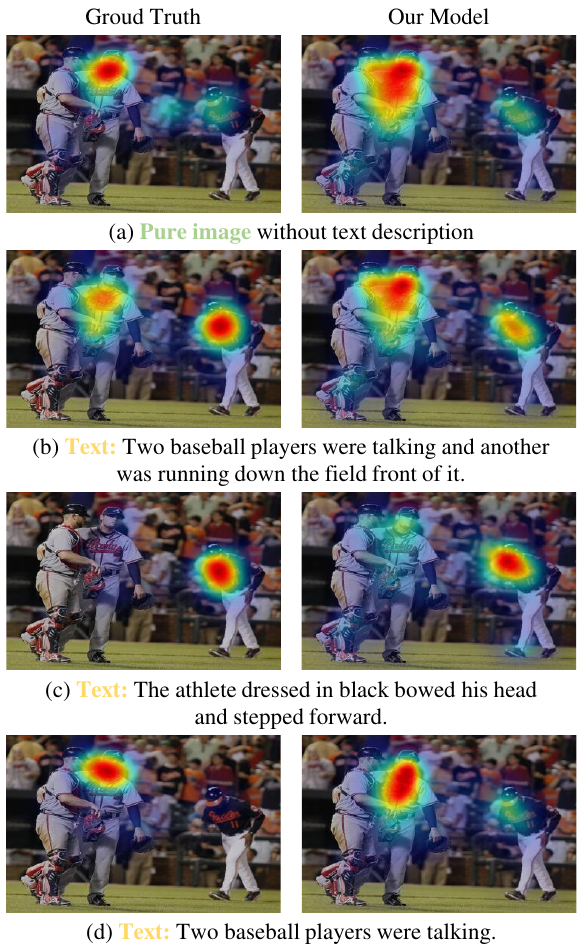}
\vspace{-1em}
\caption{The $1_{\text{st}}$ column: heatmaps of the human gaze on the original image, and the images with three different text descriptions. The $2_{\text{nd}}$ column: corresponding prediction results of our model.}
\label{heat}
\vspace{-1em}
\end{figure}

\section{Introduction}
\label{intro}
\IEEEPARstart{H}{uman} vision has the ability to select informative and conspicuous regions from external visual stimuli and attend to them, which is well known as the visual attention mechanism \cite{duan2022saliency}, \cite{borji2012state}. Human vision attention can be categorized into two functions including scene-driven bottom-up (BU) and expectation-driven top-down (TD) \cite{borji2012state}, and many studies have demonstrated that eye movements are driven by the joint influence of BU and TD attention \cite{katsuki2014bottom}. Visual attention analysis and prediction have been important tasks in multimedia and computer vision research for a long time, since they can provide new insights into the mechanisms of human attention \cite{ren2022children}, \cite{duan2019visual}, and contribute to many multimedia applications \cite{duan2018learning, duan2019dataset, min2024perceptual, duan2024quick, duan2022confusing, zhu2023audio, yang2019sgdnet} as well as various computer vision tasks \cite{zhu2021viewing, fang2020identifying, cao2023subjective, cao2023attention, gao2022image, gao2023blind, gao2022image1, yang2021progressive, cong2018review}.

Human visual attention is a complex system, thus, the analysis and prediction for it have explored the scene-driven bottom-up visual attention problem, and many corresponding saliency databases have been constructed, such as SALICON \cite{SALICON_database}, MIT1003 \cite{MIT1003}, MIT300 \cite{MIT300}, and CAT2000 \cite{CAT2000}, which are all pure image saliency databases. Based on these databases, many saliency prediction models have been proposed, which mainly include traditional handcrafted feature-based methods and deep neural network-based (DNN-based) methods. Traditional saliency prediction methods extract low-level features such as color, contrast, and semantic concepts, \textit{etc.}, and integrate these features to generate saliency maps \cite{GBVS, CA, 2013Saliency_12}. 
With the development of deep learning, many DNN-based models have also been proposed for predicting image saliency \cite{kummerer2014deep_13, SALICON, jetley2016end_15, ML-Net, che2021adversarial, zhang2020learning, yang2019dilated}. The databases and models mentioned above mainly study the scene-driven bottom-up visual attention problem, \textit{i.e.,} the pure image saliency prediction task, however, the expectation-driven top-down visual attention task has been rarely studied.

In daily life, images are often accompanied by text descriptions, such as image captions, subtitles, audio commentary, \textit{etc.} Since these text descriptions are strong expectation guidance, it is intuitive that the human visual attention to the corresponding images will be influenced by these expectations through the top-down mechanism.

As shown in the first column of Fig. \ref{heat}, when viewing the original pure image without any text description, human attention is highly attracted to the ``baseball player", however, when viewing the image accompanied by different text descriptions, the human gaze is significantly changed according to the context. Thus, it is obvious that text descriptions can significantly influence the corresponding visual attention to visual stimuli. However, to the best of our knowledge, most of current saliency models, either early hand-crafted models or recent DNN models, are not able to predict the corresponding visual saliency according to different text descriptions. Therefore, it is important to investigate new robust approaches to effectively predict human visual attention in scenes with text descriptions.

In this work, we aim to thoroughly analyze human visual attention behavior under the influence of various text descriptions and build an accurate saliency prediction model for text-guided conditions. To achieve this objective, we are facing the following research challenges.

\textbf{(i)} Building a database for text-image saliency. Although there are many publicly available image saliency databases, such as SALICON \cite{SALICON_database} and MIT300 \cite{MIT300}, they are all pure images without text descriptions. In addition, some other databases, such as MSCOCO \cite{MSCOCO} and Flickr30k database \cite{Flickr30k}, contain images with text descriptions, but there is no corresponding ground truth visual attention data.

\textbf{(ii)} Understanding the effect of various text descriptions on visual saliency. As a common observation, without text descriptions, the visual attention mechanism will make people pay more attention to the informative and conspicuous areas of an image. Moreover, a text description can influence the visual attention on the image \cite{sun2023influence}. However, whether and how different text descriptions of one image influence the corresponding visual attention is still unknown.

\textbf{(iii)} Modeling text-image saliency. Since images are usually accompanied by texts in daily life and different descriptions cause different influences on the corresponding visual attention. It is necessary and significant to study how to integrate both image features and text features, and jointly exploit these two parts of information to build an accurate text-image saliency model.

\begin{figure}[!t]
\centering
\includegraphics[width=3in]{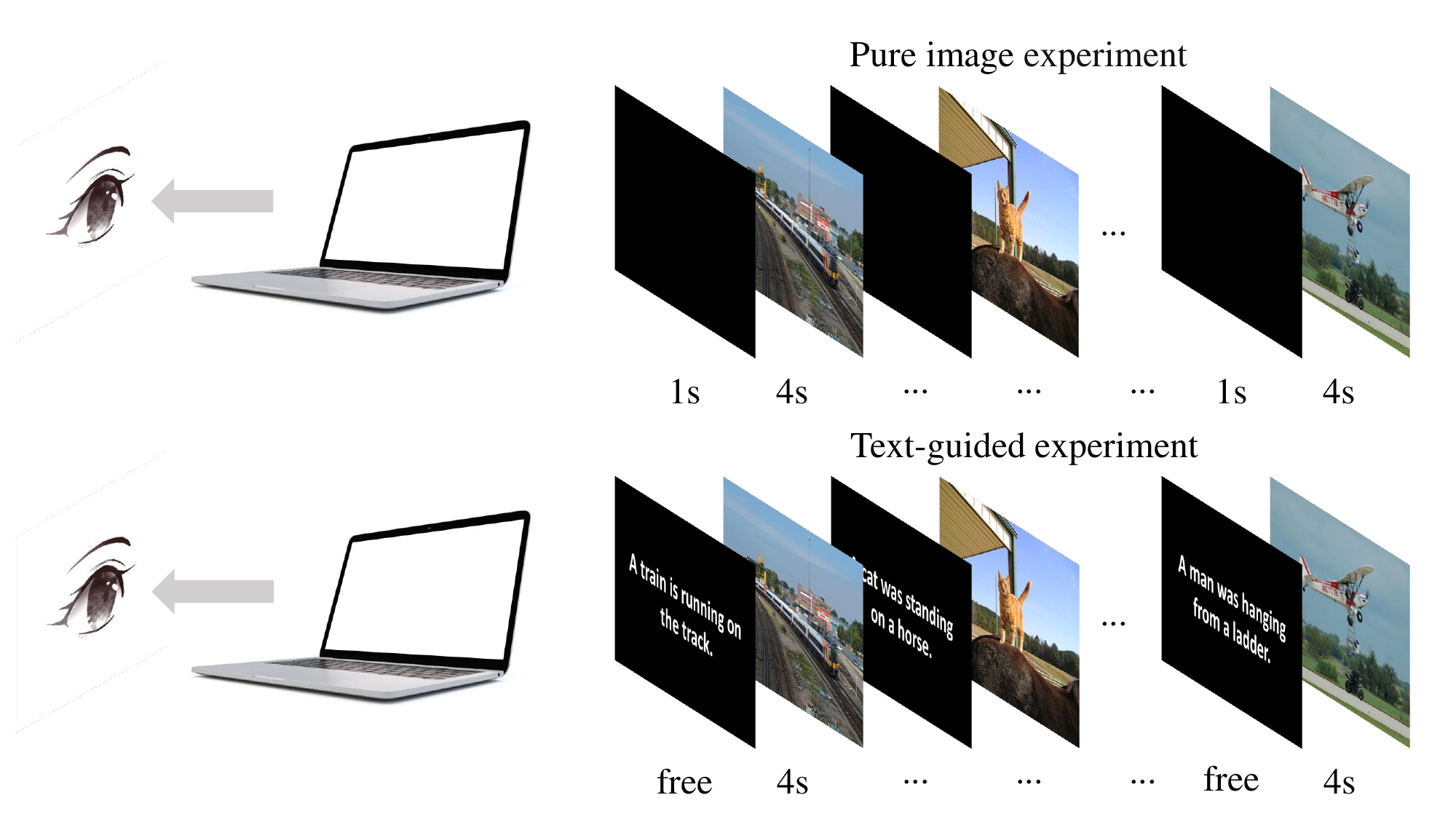}
\vspace{-1em}
\caption{Schematic diagram of our eye-tracking experiment. Comparison between pure image condition and text-guided condition is shown.}
\label{expsetup}
\vspace{-1em}
\end{figure}

\begin{figure}[!t]
\centering
\includegraphics[width=3.4in]{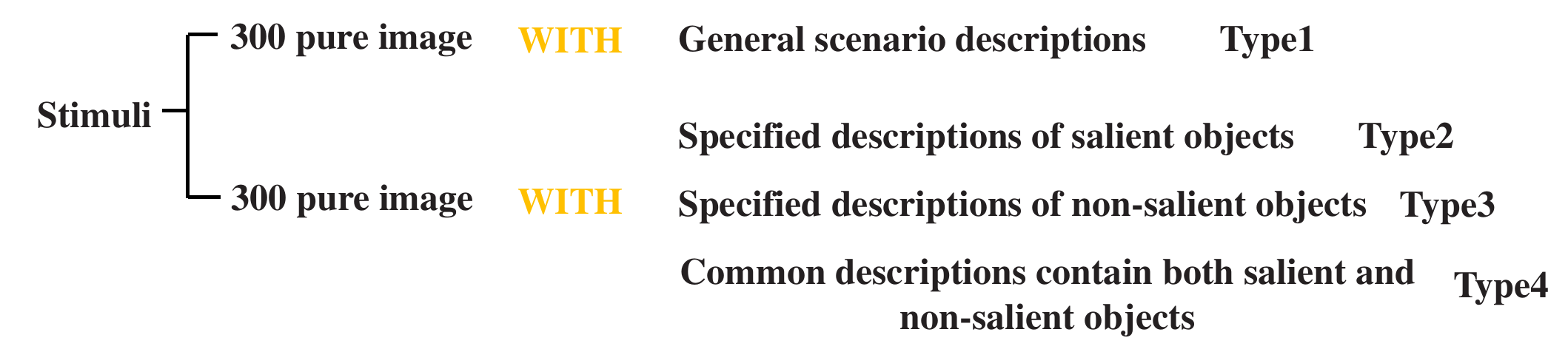}
\vspace{-1em}
\caption{Classification of the attributes of the texts.}
\label{Classification}
\vspace{-1em}
\end{figure}

In this work, we first construct a text-guided image saliency database, termed SJTU-TIS.
Specifically, as shown in Fig. \ref{expsetup}, in order to investigate whether a text description can influence visual attention on an image, the eye tracking experiment is conducted under two conditions including a pure image condition and a text-guided condition.
Moreover, as shown in Fig. \ref{Classification}, in order to investigate how different text descriptions influence the corresponding visual attention, the collected 600 images are divided into two parts, including 300 images with general scenario descriptions and 300 images with three different types of object descriptions. Overall, our SJTU-TIS database has 600 images and 1200 text descriptions, which results in 1200 text-image pairs. To better predict the visual attention influenced by a text description, we propose a novel text-guided saliency prediction model (TGSal) that can be used to predict visual saliency under both pure image and text-guided conditions. Specifically, we encode the text features and image features into the embedding space and inject the text features to image features step by step at the decoding end to get the final prediction results. Experimental results demonstrate the effectiveness of the proposed text feature fusion modules on the text-guided saliency prediction task, and our proposed TGSal achieves the best performance on both the pure image saliency databases and our constructed SJTU-TIS database. The contributions are summarized as follows:

\begin{itemize}
\item{We build a new text-guided image saliency database, termed SJTU-TIS, which aims to study the influence of different text descriptions on the corresponding visual saliency of an image.}
\item{We analyze the effects of different text descriptions on visual attention, and indicates that image saliency is significantly influenced by a text description, and different text descriptions for the same image may have different impacts on the corresponding visual attention.}
\item{We validate the performance of state-of-the-art unimodal saliency prediction models and multimodal text-image pretraining methods on our SJTU-TIS database, and establish a benchmark.}
\item{A prediction model for text-image visual saliency is proposed, which achieves the best performance under both pure image and text-guided conditions compared to benchmark methods.}
\end{itemize}

The rest of the paper is organized as follows. Section \ref{Related Work} introduces related works. Then we introduce the construction procedure and the analysis of the SJTU-TIS database in Section \ref{Saliency in Text-image database}. Section \ref{Proposed Method} introduces our proposed TGSal model in detail. Section \ref{Experimental Setup and Results} describes the experimental settings and the experimental results. Section \ref{Conclusion} concludes the whole paper.

\section{Related Work}
\label{Related Work}
\subsection{Eye-tracking Databases}
\subsubsection{Traditional Saliency Databases}
In order to understand and model visual attention behavior, many eye-tracking databases have been constructed under the free-viewing task. MIT1003 \cite{MIT1003} is a large-scale saliency database that contains 1003 images coming from Flickr and LabelMe. MIT300 \cite{MIT300} and CAT2000 \cite{CAT2000} are two widely used benchmark databases, which contain 300 and 2000 test images respectively. SALICON \cite{SALICON_database} is currently the largest crowd-sourced saliency database, which contains 10000 training images, 5000 validation images, and 5000 test images, which is widely adopted for pretraining saliency prediction models. The saliency data is collected through mouse tracking using Amazon Mechanical Turk (AMT). 

\subsubsection{Text-image Saliency Database}
In our previous work \cite{sun2023influence}, we have established a text-image saliency database.
As shown in Table \ref{3different_dataset}, the previous dataset mainly aims to investigate whether the text description can influence the saliency, thus only contains one pure image saliency map and one text-image saliency map for one image. Since the previous work has validated the influence of texts on image saliency, our new dataset SJTU-TIS further aims to investigate whether and how different texts can influence the image saliency, which has not been involved and studied in the previous work. Moreover, the current database is a rebuilt dataset with more images. 
To investigate the influence of different texts on image saliency, we remove some images and keep about 200 images from the previous dataset. In summary, the SJTU-TIS database contains 300 pure image saliency maps and 300 text-image saliency maps for a group of 300 images, and includes 300 pure image saliency maps and 300$\times$3 text-image saliency maps for another group of 300 images.

\vspace{-0.5em}
\subsection{Saliency Prediction Models}
\subsubsection{Classical Models}
Most traditional methods model visual saliency based on the bottom-up mechanism, which generally extract simple low-level feature maps, such as intensity, color, direction, \textit{etc.}, and integrate them to generate saliency maps. Itti \textit{et al.} \cite{IT} considered underlying features on multiple scales to predict saliency maps. Harel \textit{et al.} \cite{GBVS} introduced a graph-based saliency model, which defined Markov chains on various image maps and regarded the balanced distribution of map locations as activation values and saliency values. 
Many other classical methods such as AIM \cite{AIM}, SMVJ \cite{SMVJ}, CovSal \cite{CovSal}, SeR \cite{SeR}, HFT \cite{HFT}, \textit{etc.}, are also commonly used saliency prediction models.

\subsubsection{Deep Models}
With the development of deep neural networks (DNN), saliency prediction tasks have made significant progress in recent years \cite{kruthiventi2017deepfix_23, tu2022end_24}. 
Huang \textit{et al.} \cite{SALICON} used VGG as the backbone and proposed a two-stream network to extract coarse features and fine features to calculate saliency maps. Cornia \textit{et al.} \cite{SAM} proposed an Attentive ConvLSTM, which focuses on different spatial positions of a bunch of features to enhance prediction. Pan \textit{et al.} \cite{SalGAN} proposed the generative adversarial network (GAN) to calculate the saliency map. Che \textit{et al.} \cite{GazeGAN} studied the influence of transformation on visual attention and proposed a GazeGAN model based on the U-Net for saliency prediction. Duan \textit{et al.} \cite{duan2022saliency} proposed a vector quantized saliency prediction method and generalized it for AR saliency prediction. These top-down-based saliency prediction models have been widely used in various research fields in recent years \cite{borji2019saliency_32}.

\subsection{Text-image Pretraining and Text-to-image Synthesis}
In recent years, many large-scale text-image pretrained basic models have been proposed. Li \textit{et al.} \cite{li2021align_33} proposed a contrastive loss to align the image and text representations before fusing them through cross-modal attention, which enables more grounded vision and language representation learning. Radford \textit{et al.} \cite{radford2021learning_34} proposed the CLIP model, which takes texts and images as input, performs cosine similarity calculation on encoder results, and obtains classification results. The development of text-image pretraining models has also promoted the development of text-to-image synthesis. 
The Latent Diffusion Model (LDM) \cite{rombach2022high_39} performs the processing in the low-dimensional latent space rather than the pixel space. Li \textit{et al.} \cite{li2023gligen_40} froze the original LDM structure and added a new trainable self-attention module to increase input to generate a graph ``under specific conditions", which injected and learned new information while retaining the original pretraining information. 
Kang \textit{et al.} \cite{kang2023scaling_42} proposed a multi-scale GAN network (GigaGan) to generate images and found that interlacing self-attention (image only) and cross-attention (text-image) with convolutional layers can improve performance. This network has a faster image generation speed and weaker visual quality than the diffusion network.

\section{SJTU-TIS Database}
\label{Saliency in Text-image database}
Due to the absence of a text-image saliency database, in this paper, we construct the first text-guided image saliency database, denoted as the SJTU-TIS database. We first select 600 images from MSCOCO \cite{MSCOCO} and Flickr30k \cite{Flickr30k} with the corresponding text descriptions. Then, a subjective experiment is conducted to obtain the eye movement data, which is processed to obtain the visual attention map of the SJTU-TIS database.

\begin{table}[t]

    \centering
    \caption{Comparison between the current database (SJTU-TIS) and the previous database (ISCAS 2023 \cite{sun2023influence}).}
    \vspace{-1em}
    \label{3different_dataset}
    \setstretch{1}
    \fontsize{12}{13.8}\selectfont
    \resizebox{0.48\textwidth}{!}{
    \begin{tabular}{m{3.7cm}|m{4cm}|m{4cm}}
    \toprule[1.5pt]
                & Journal\enspace version (SJTU-TIS)      & Conference\enspace version (ISCAS 2023 \cite{sun2023influence})  \\ \hline
    Objective                        & Investigate\enspace whether and\enspace how\enspace \textbf{different} texts\enspace can influence\enspace the image\enspace saliency & Investigate whether the text can influence the image saliency \\ \hline
    Number\enspace of\enspace images                      & 600                         & 300         \\ \hline    
    Number\enspace of\enspace texts                       & 1200                        & 300         \\ \hline
    Number\enspace of\enspace saliency maps               & 1800                        & 600         \\ \hline
    Relationship\enspace between image\enspace and\enspace text  & One\enspace to\enspace one\enspace or\enspace one\enspace to many   & One\enspace to\enspace one       \\ 
    \bottomrule[1.5pt]
    \end{tabular}
    }
    \vspace{-1em}
\end{table}

\begin{figure}[!t]
\centering
\includegraphics[width=2.5in]{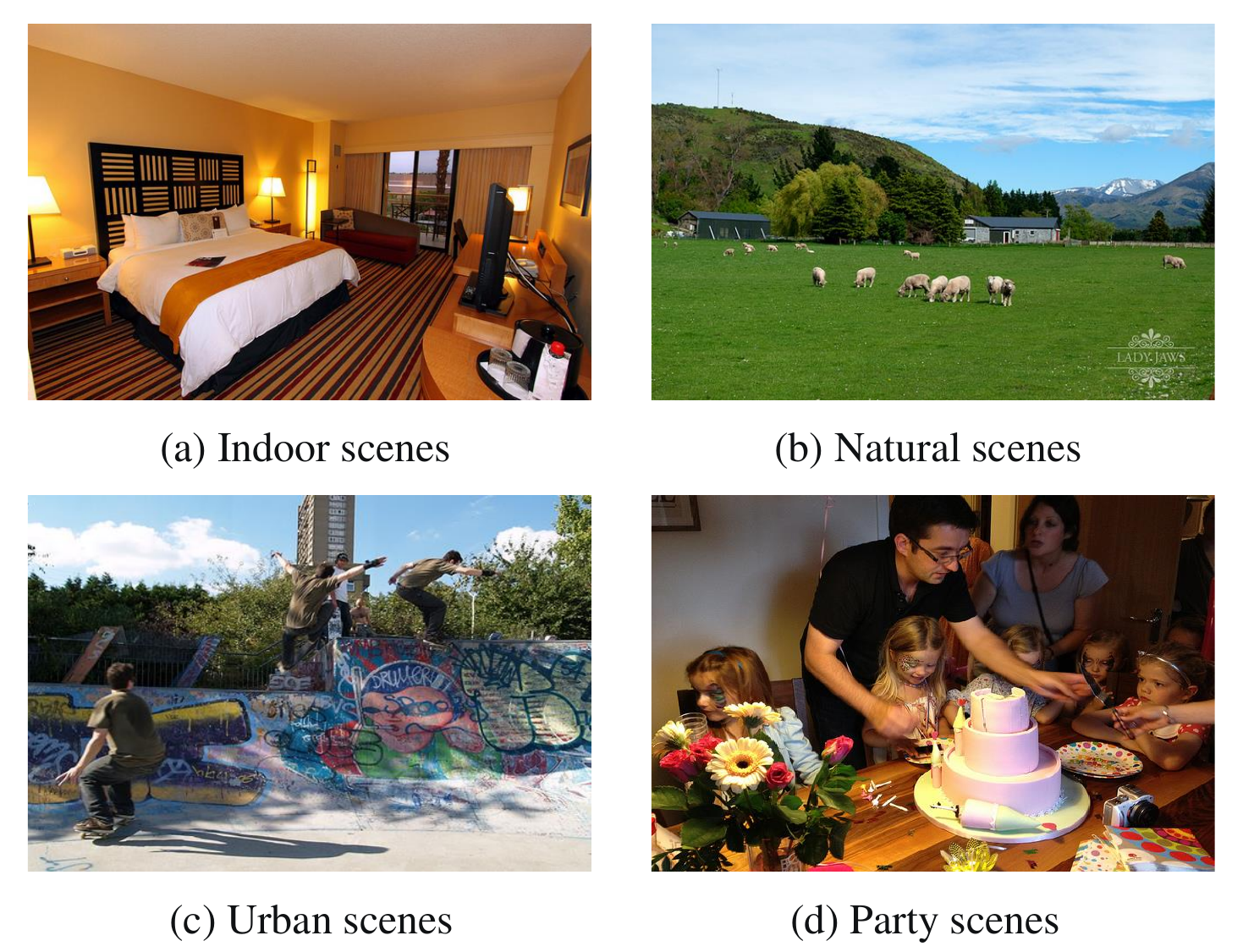}
\vspace{-1em}
\caption{Examples of the collected different scenes. (a) Indoor scenes. (b) Natural scenes. (c) Urban scenes. (d) Party scenes.}
\label{scenes}
\vspace{-1em}
\end{figure}

\vspace{-0.5em}
\subsection{Text-image Pair Collection}
 To collect images with diverse scenes, we first selected 4 scenarios from the MSCOCO database \cite{MSCOCO} and the Flickr30k database \cite{Flickr30k}, including indoor scenes, natural scenes, urban scenes, and party scenes, as shown in Fig. \ref{scenes}.  
 Although each image in the MSCOCO database \cite{MSCOCO} and the Flickr30k database \cite{Flickr30k} has multiple corresponding text descriptions, the semantic meanings of these text descriptions are similar, which does not meet the study requirements. Since our objective is to study the visual saliency of an image under different descriptions, we manually modified these text descriptions to four conditions as shown in Fig. \ref{Classification}. For natural scenes, since they usually do not include salient or non-salient objects, we only produce general scenario descriptions. For other scenes that have salient objects and non-salient objects, we produce three text descriptions for each image, including specified descriptions for salient objects, specified descriptions for non-salient objects, and common descriptions for both salient and non-salient objects. Therefore, our SJTU-TIS database contains 600 images and 1200 text descriptions, which results in 1200 text-image pairs (300+300$\times$3) in total. Moreover, the length of all sentences is limited to 5-25 words in order to simulate practical application scenarios and facilitate the subjective experimental setting.

\begin{figure}[!t]
\vspace{-0.5em}
\centering
\includegraphics[width=3.4in]{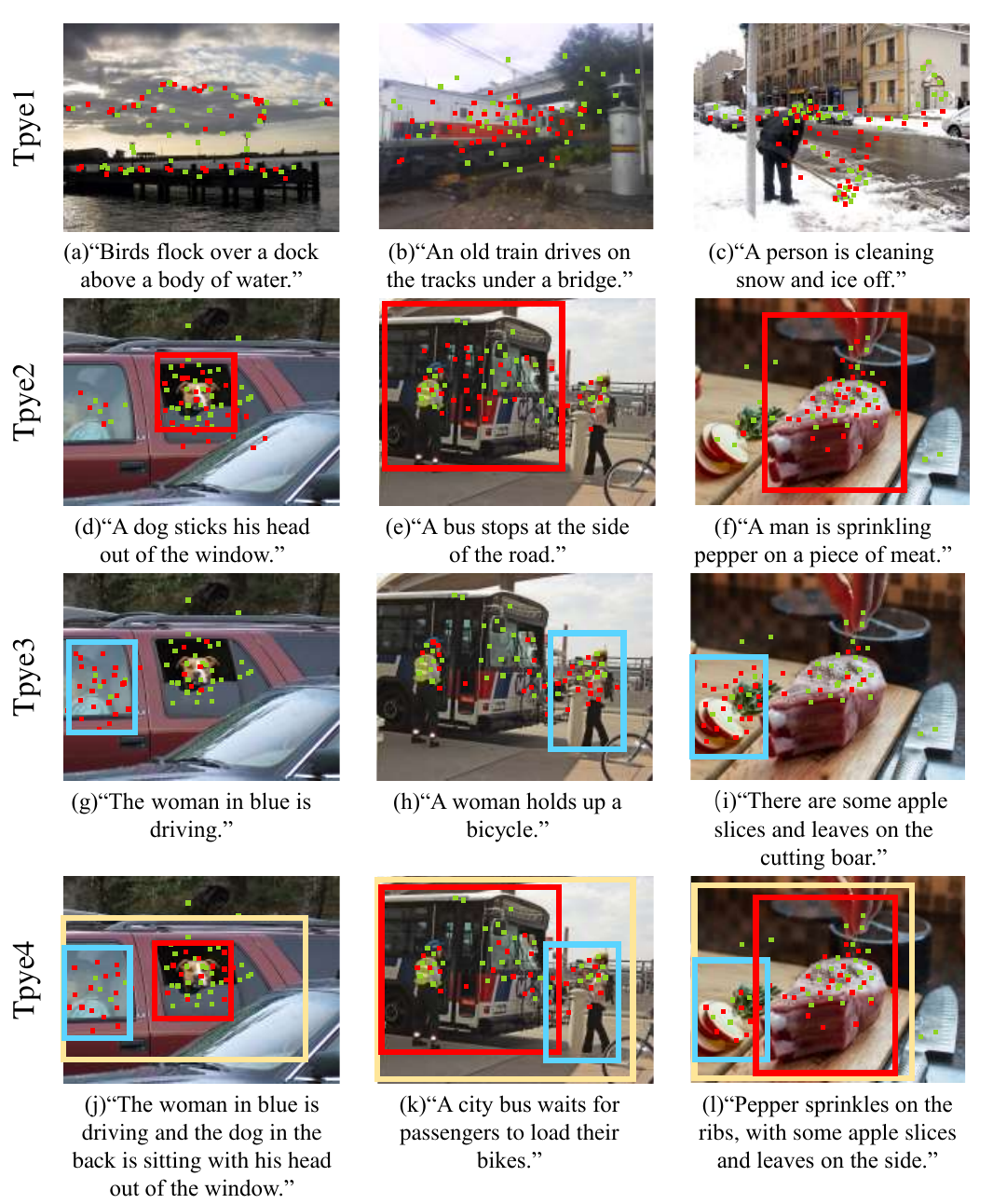}
\vspace{-1em}
\caption{An illustration of the example images and the corresponding fixation schematic map under four different text-description conditions given in Fig. \ref{Classification}. Red boxes represent salient objects, blue boxes represent non-salient objects, and yellow boxes represent common descriptions containing both salient and non-salient objects. Red points: text-guided condition. Green points: pure image condition.}
\label{fixation_and_describe}
\vspace{-1.5em}
\end{figure}

 Fig. \ref{fixation_and_describe} shows some representative images from the SJTU-TIS database. The first row represents the first group of images (Type1), and the second to fourth rows demonstrate the second group of images (Type2, 3, and 4). The red rectangular box represents the specified descriptions of salient objects, the blue rectangular box represents the specified descriptions of non-salient objects, and the yellow rectangular box represents the common descriptions containing both salient and non-salient objects. It is obvious that different text descriptions correspond to different image areas, which may significantly influence the corresponding visual attention.

\vspace{-1em}
\subsection{Subjective Eye-tracking Experiment}
We conducted a subjective eye-tracking experiment to obtain the visual attention maps of the images in the SJTU-TIS database using the Tobii Pro X3-120 eye-tracker \cite{tobii}. Tobii Pro X3-120 is an ultra-thin and lightweight portable eye tracker that can be directly combined with various types of screens. The eye-tracking device can be directly connected to a computer through USB. We recorded eye movements at a sampling rate of 120Hz. The resolution of the screen is 1920$\times$1080, and the screen resolution can be freely adjusted. The acceptive size of the head-sport box is 50cm in width, 40cm in height, and 80cm in length, with a tracking distance of 50-90cm.

We designed the experimental process using the software provided by the Tobii Pro X3-120 eye tracker. The eye-tracking experiment is divided into 5 sessions. We set Type1 to Type4 into four sessions (each with 300 text-image pairs), and set all pure images into one session, which is the fifth session (containing 600 pure images). All images were displayed at their raw resolutions. To avoid fatigue, subjects had a break time after viewing every 100 images. As shown in Fig. \ref{expsetup}, during the experiment, each image was displayed for 4 seconds. In the pure image condition, each image stimuli was followed by a 1-second black screen interval. In the text-guided condition, there was a text-description viewing while before displaying each image. The duration of viewing the text was controlled by the participants through the mouse click, \textit{i.e.}, after reading and understanding the text, they can click the mouse to view the corresponding image. The display time of each image is still 4 seconds. After four seconds, the next text description was automatically shown. The distance between the subjects and the eye tracker was roughly maintained between the range of 2 and 2.3 feet during each session. The eye-tracking experiment was conducted in a quiet room.

A total of 60 subjects were recruited to participate in the experiment. Since subjects may remember an image if they have seen this image before, which may affect the reliability of the collected eye movement data, we ensured that each subject only viewed each image once.
Specifically, the 60 subjects were divided into 4 groups with 15 subjects in each group, and the first group participated in the pure image experiment, the second group participated in the ``Type1'' and ``Type2'' text-guided eye-tracking experiment, the third and the fourth group participated in the ``Type3'' and ``Type4'' text-guided eye-tracking experiment, respectively. Therefore, each pure image or text-image pair was viewed by 15 participants without repeating. 
The order of these images was intentionally randomized within each session to reduce potential bias in the experimental process. Furthermore, to maintain participant engagement and concentration, each session was split into several parts, of which each part included a group of 100 images. Participants were given a rest period after each part in the session, ensuring sustained attention and focus throughout the experiment.
Before the experiment, each subject first read an instruction explaining the experimental process and experimental requirements, and then experienced a brief training session to be familiar with the experimental procedure. All subjects had normal or corrected normal vision.

\vspace{-1em}
\subsection{Data Processing and Analysis}

\begin{figure*}[!t]
\centering
\includegraphics[width=6in]{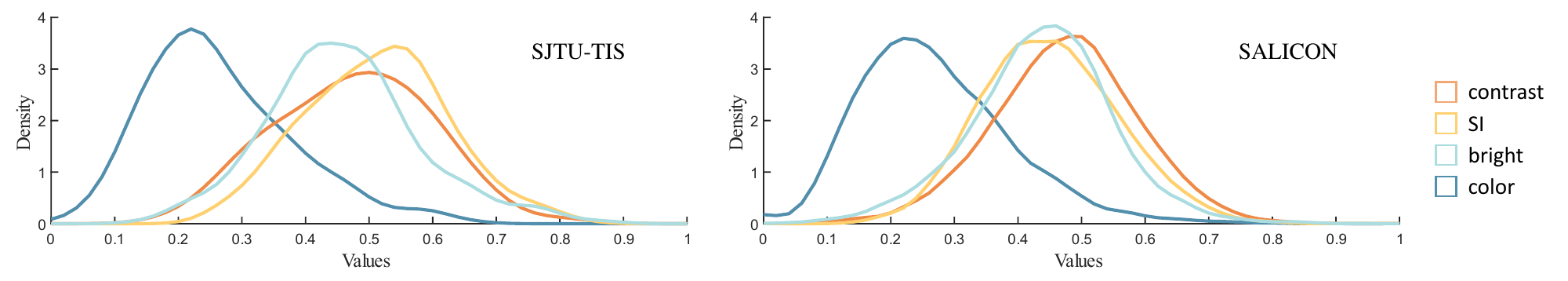}
\vspace{-1em}
\caption{Distribution comparisons of the image attributes between two databases: SJTU-TIS, SALICON \cite{SALICON_database}.}
\label{feature_database}
\vspace{-1em}
\end{figure*}

\subsubsection{Image Attribute Analysis}
We analyze four image attributes, including contrast, colorfulness, spatial information, and brightness to characterize the content diversity of the images in the SJTU-TIS database. The image attributes of the SJTU-TIS and SALICON databases \cite{SALICON_database} are shown in Fig. \ref{feature_database}. It can be observed that our SJTU-TIS database covers a wide range of content diversity.

\subsubsection{Eye-tracking Data Processing and Analysis}
If the overall sampling rate of the eye movement is less than 90\%, the data and the subject will be regarded as outlier. None of the 60 subjects is identified as outlier and removed from the experiment. We first overlay all fixation points of one image fixated by all viewers into one map to generate the fixation map of this image. Then the fixation map is smoothed with a 1$^{\circ}$ Gaussian kernel to obtain a continuous fixation density map (visual attention map).

Fig. \ref{fixation_and_describe} shows some schematic diagrams of the fixation maps. We use green points to represent the fixations obtained under the pure image condition and red points to represent the fixations obtained under the text-guided condition. The first line represents the general scenario descriptions (Type1), while the second to fourth rows represent three different descriptions of the same image (Type2-Type4). 
It can be observed that for Type1, there is not much difference between the distributions of red and green points, which are both relatively uniform. For Type2, under the specified descriptions of salient objects, red points are more concentrated on the described object compared to green points, such as the meat in the third image. For Type3, when describing non-salient objects, there is a significant difference between the red points and green points. Green points are rarely distributed on the described non-salient object, while red points are concentrated on the described object, such as the apple slices in the third image. For Type4, under the condition of common descriptions containing both salient and non-salient objects, the red points will partially shift to non-salient objects, while most of them concentrated on salient objects.

\begin{figure}[!t]
\centering
\includegraphics[width=3in]{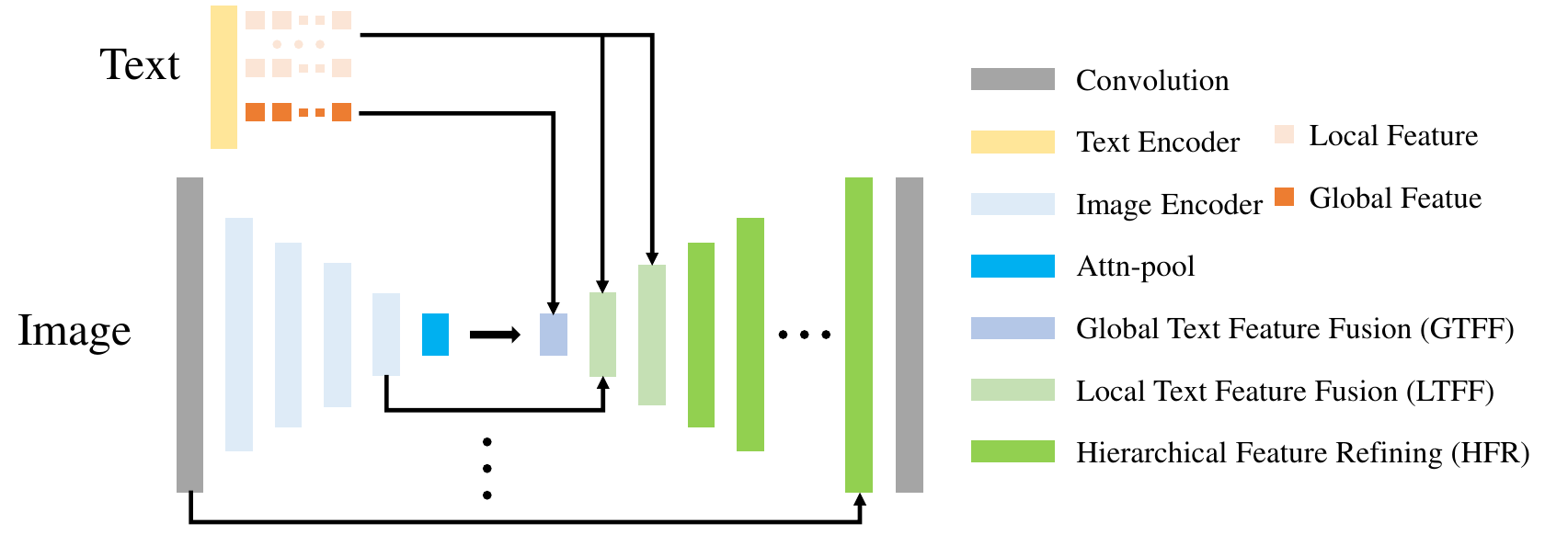}
\vspace{-1em}
\caption{An overview of the structural diagram of the proposed method.}
\label{UNet}
\vspace{-1em}
\end{figure}

\begin{figure*}[!t]
\centering
\includegraphics[width=6in]{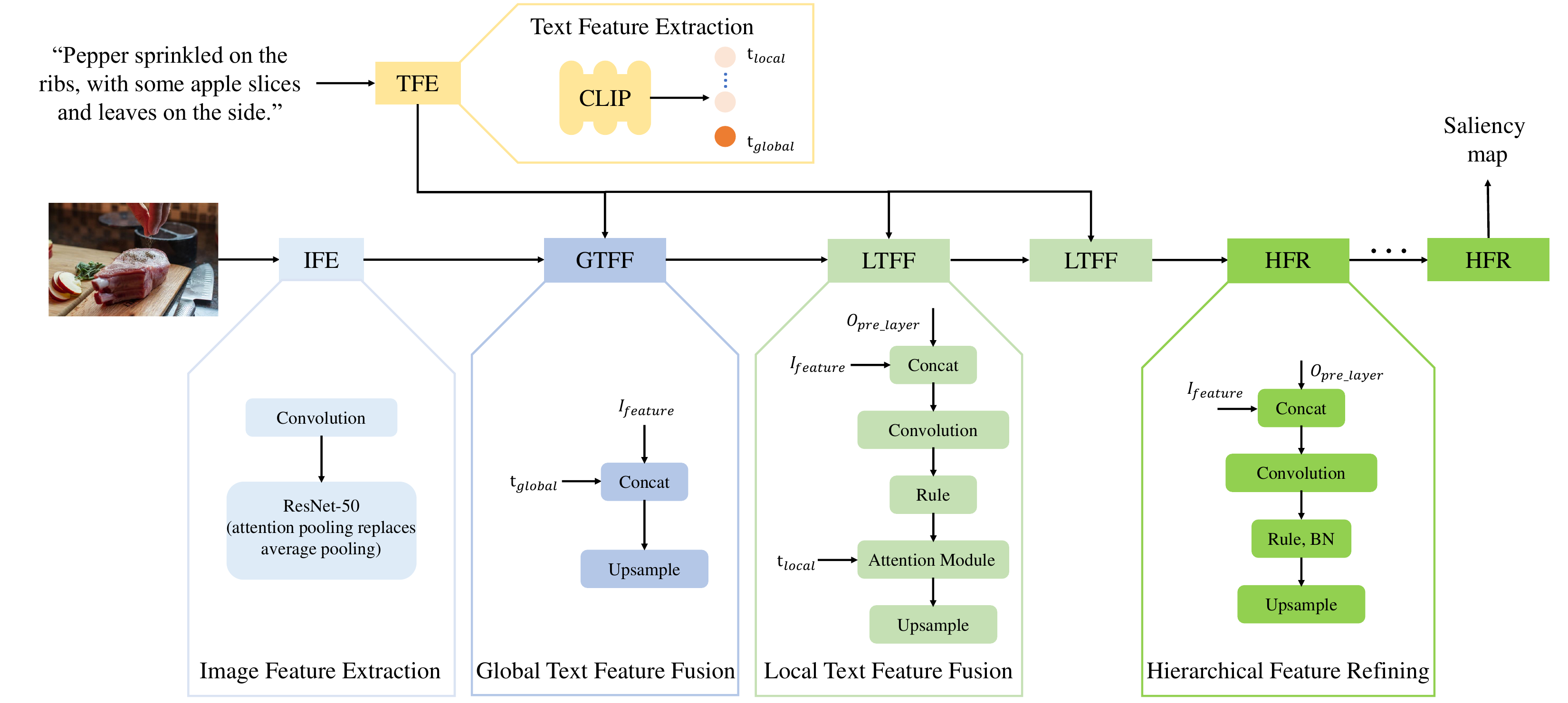}
\vspace{-1em}
\caption{The detailed architecture of the proposed method, which is composed of five modules. (a) The image feature extraction module (\textcolor[RGB]{222,235,247}{IFE}) extracts image features using ResNet-50. (b) The text feature extraction module (\textcolor[RGB]{255,230,153}{TFE}) extracts text features using the CLIP text encoder \cite{radford2021learning_34}. (c) The global text feature fusion module (\textcolor[RGB]{180,199,231}{GTFF}) fuses the global text feature and the image feature. (d) The local text feature fusion module (\textcolor[RGB]{197,224,180}{LTFF}) fuses the local text feature and image feature. (e) The hierarchical feature refining module (\textcolor[RGB]{146,208,80}{HFR}) refines the integrated image and text features, and outputs the predicted saliency map.}
\label{model}
\vspace{-1em}
\end{figure*}

\begin{figure}[!t]
\centering
\includegraphics[width=3in]{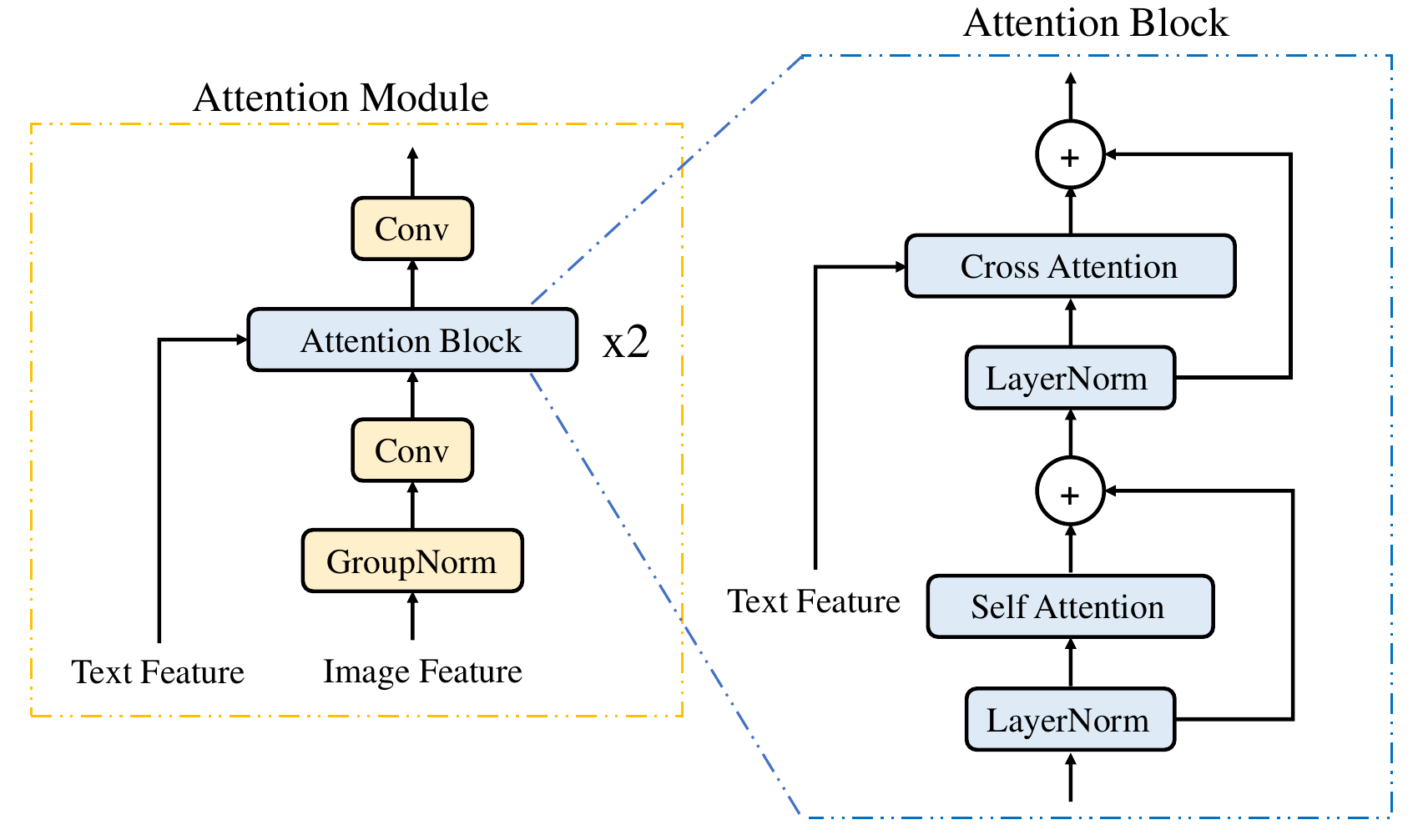}
\vspace{-1em}
\caption{The framework of attention module in the local text feature fusion module.}
\label{Attention}
\vspace{-1.5em}
\end{figure}

\section{Proposed Method}
\label{Proposed Method}

In this section, we describe the architecture of our proposed text-guided visual saliency prediction (TGSal) model in detail. The overall structure of the model is shown in Fig. \ref{UNet}. The proposed TGSal model is composed of five main modules including the image feature extraction module (IFE), the text feature extraction module (TFE), the global text feature fusion module (GTFF), the local text feature fusion module (LTFF), and the hierarchical feature refining module (HFR), as described in Section \ref{Feature Extraction} and Section \ref{Feature Fusion}. Firstly, the image feature extraction module and the text feature extraction module extract features of images and the corresponding text descriptions, respectively. Then, the extracted global text features are fed into the global feature fusion module to fuse with the image features, and the extracted local text features are fused with image features at multiple scales. Finally, the hierarchical feature refining module refines the integrated image and text features, and outputs the predicted saliency map. The detailed architecture of the proposed TGSal model is shown in Fig. \ref{model}. Two loss functions are adopted to optimize the whole network as described in Section \ref{Loss Function}.
 
\vspace{-1em}
\subsection{Feature Extraction}
\label{Feature Extraction}
As shown in Fig. \ref{model}, we use ResNet-50 as the image feature extraction (IFE) module to extract image features, which can provide multi-scale features for the hierarchical decoding process.
Due to the different resolutions of the images in the SJTU-TIS database, we unify the size of all images into 224$\times$224 before feeding into the first convolutional layer. Cornia \textit{et al.} \cite{SAM} mentioned that the rescaling caused by max-pooling deteriorates the performance of saliency prediction, therefore, we change the final pooling layer to the attention pooling. Given an image $I$, the IFE module is formulated as:
\begin{equation}
F_i = [I_1, I_2, ..., I_n, I_{\text{bottleneck}}],
\end{equation}
where $F_i$ indicates the extracted image features, $I_i, i={1, 2, ..., n}$ represents hierarchical image features of the image encoder, and $I_{\text{bottleneck}}$ means the extracted bottleneck image feature.

For text feature extraction, since it is hard to train an effective text encoder that can well align with the image encoder from scratch, inspired by Kang \textit{et al.} \cite{kang2023scaling_42}, we choose a pretrained text encoder, \textit{i.e.}, CLIP text encoder, to obtain global and local semantic information for text descriptions. Given a text description represented in the following form:
\begin{equation}
 \text{Text} : t = [t_{1},t_{2},...,t_{L}],
\end{equation}
where $L$ is the length of the text, each component $t_{i}$ of $t$ is the $i_{th}$ word in the sentence. We obtain the text feature vectors through the TFE module demonstrated in Fig. \ref{model} as:
\begin{equation}
F_{t}=f_{\text{text}}(t),
\label{eq2}
\end{equation}
where $f_{\text{text}}$ indicates the text feature extraction module.
The obtained text features contain two components including a global text feature vector $T_{\text{global}}$ and a local text feature vector $T_{\text{local}}$:
\begin{equation}
F_{t} = (T_{\text{global}},T_{\text{local}}),
\end{equation}
\begin{equation}
\text{Text}: T_{\text{global}}=[x_{1},x_{2},...,x_{L}],
\end{equation}
\begin{equation}
\text{Text}: T_{\text{local}} =[w_{1},w_{2},...,w_{L}],
\end{equation}
where $L$ is the length of the text, the maximum is 77, $x_{1}$ to $x_{L}$ represents the global semantic information, $w_{1}$ to $w_{L}$ represents the local semantic information, and $w_{q}$ indicates the contextual text features of the $q_{th}$ word in the text.

\subsection{Feature Fusion}
\label{Feature Fusion}

For the global text feature $T_{\text{global}}$ extracted in Section \ref{Feature Extraction}, we concatenate it with the image feature at the bottleneck layer. Then we adopt an upsample layer to perform global image and text feature integration and dimension expansion. The GTFF module can be described as:
\begin{equation}
 F_{\text{global}} = g_{\text{up}}(g_{\text{concat}}(I_{\text{bottleneck}},T_{\text{global}})),
\end{equation}
where $T_{\text{global}}$ represents the global features of the text, $I_{\text{bottleneck}}
$ represents the corresponding bottleneck image feature extracted by the IFE module, $g_{\text{up}}$ represents the upsample manipulation, and $g_{\text{concat}}$ represents the feature concatenation operation.

For the local text feature $T_{\text{local}}$ extracted in Section \ref{Feature Extraction}, it is layered and injected together with the multi-scale image features from the IFE module. 
The LTFF module is formulated as:
\begin{equation}
F_{\text{local}} = g_{\text{up}}(g_{\text{attn}}(g_{\text{conv}}(g_{\text{concat}}(I_{i}^{\text{up}}, I_{i})),T_{\text{local}})),
\end{equation}
where $g_{\text{up}}$, $g_{\text{attn}}$, $g_{\text{conv}}$, and $g_{\text{concat}}$ represent the upsample layer, attention module, convolutional layer, and feature concatenation, respectively, $T_{\text{local}}$ represents the local text features, and $I_{i}^{\text{up}}$ represents the upsampled image feature vector from the previous decoder level, and $I_{i}$ means the image feature vector from the IFE module through the skip connection.
Specifically, the detailed structure of the attention module is shown in Fig. \ref{Attention}. Given an input feature $v=[v_{1}, ..., v_{M}]$, where $M$ is the length of the feature vector output from the previous module, the attention module can be described as:
\begin{equation}
v = v + \text{SelfAttn}(v),
\end{equation}
\begin{equation}
v = v + \text{CrossAttn}(v,T_{\text{local}}).
\end{equation}
We find that adding some attention layers to each layer is beneficial to the performance, but excessive attention layers can cause overfitting on the SJTU-TIS database. The detailed discussion of the structure of the LTFF module will be given in Section \ref{Experimental Setup and Results}.

Since adding attention layers to high resolution features can dramatically increase the GPU memory, we do not include the attention module in the HFR module. The detailed structure of the HFR module is shown in Fig. \ref{model}, which can be formulated as:
\begin{equation}
F_{{\text{out}}} = g_{\text{up}}(g_{\text{conv}}(g_{\text{concat}}(I_{i}^{\text{up}}, I_{i}))),
\end{equation}
where $g_{\text{up}}$, $g_{\text{conv}}$, $g_{\text{concat}}$ represent the upsampling layer, convolutional layer, and image feature concatenation, respectively, $I_{i}^{\text{up}}$ represents the upsampled image feature vector from the previous decoder level, $I_{i}$ means the image feature vector from the IFE module through the skip connection.

\vspace{-1em}
\subsection{Loss Function}
\label{Loss Function}
Pan \textit{et al.} \cite{SalGAN} proposed that in the design of loss functions, relying solely on certain saliency metrics may lead to poor results for other saliency metrics that are not introduced in the loss function. In order to obtain better saliency prediction results in terms of various evaluation metrics, we introduce a loss function, which is formed by a linear combination of correlation coefficient and mean square error. We define the overall loss function as:
\begin{equation}
\mathcal{L}(\hat{y},{y_{\text{den}}})=\alpha \cdot (1-\mathcal{L}_1(\hat{y},{y_{\text{den}}}))+\beta \cdot \mathcal{L}_{2}(\hat{y},{y_{\text{den}}}),
\end{equation}
where $\hat{y}$ and {$y_{\text{den}}$} are the predicted saliency map and the corresponding ground truth fixation density map, respectively, while $\alpha$ and $\beta$ are two hyper-parameters that balance the two loss functions. $\mathcal{L}_{1}$ and $\mathcal{L}_{2}$ are the Linear Correlation Coefficient (CC) and the Mean Square Error (MSE), respectively, which are commonly used as the loss function of saliency model training.

CC treats the saliency and ground truth density maps, $\hat{y}$ and {$y_{\text{den}}$}, as random variables and measures the linear relationship between them, which can be computed as:
\begin{equation}
\mathcal{L}_{1}(\hat{y},{y_{\text{den}}})= \frac {\sigma(\hat{y},{y_{\text{den}}})}{\sigma(\hat{y})\cdot\sigma({y_{\text{den}}})},
\end{equation}
where $\sigma(\hat{y},{y_{\text{den}}})$ is the covariance of $\hat{y}$ and {$y_{\text{den}}$}, $\sigma(\hat{y})$ and $\sigma({y_{\text{den}}})$ represent the variances of $\hat{y}$ and {$y_{\text{den}}$}, respectively. 

MSE is the average of the square of the difference between $\hat{y}$ and {$y_{\text{den}}$}, which can be formulated as:
\begin{equation}
\mathcal{L}_{2}(\hat{y},{y_{\text{den}}}) = \frac {1}{N}\sum _{i=1}^{N}({y_{\text{den}}} - \hat{y})^{2},
\end{equation}
{where $N$ is the number of prediction results.} Since the prediction result is closer to the ground truth, the CC is larger and the MSE is smaller, we use $1-\mathcal{L}_1$ instead of $\mathcal{L}_1$ when calculating the CC loss.

\section{Experiments and Results}
\label{Experimental Setup and Results}
In this section, we first introduce our experimental settings, including the test databases, evaluation metrics, and implementation details.
Secondly, we quantitatively and qualitatively compare the proposed method with the benchmark saliency prediction models on a common saliency prediction dataset, \textit{i.e.,} SALICON, and the constructed SJTU-TIS database. Then, we introduce our ablation studies to validate the effectiveness of each module of our proposed model. Finally, we compare the running speed of our proposed TGSal model with other state-of-the-art saliency models to demonstrate the superiority of our model.

\vspace{-1em}
\subsection{Experimental Setup}

\subsubsection{Datasets}
In order to understand and predict visual attention behavior, many eye-tracking databases have been constructed in recent years. In this paper, two databases are used to validate the effectiveness of the proposed TGSal model, including the largest publicly available saliency database SALICON \cite{SALICON_database} which is constructed for the general saliency prediction purpose, and the proposed SJTU-TIS database which is established for the text-guided saliency prediction purpose.

\subsubsection{Evaluation Metrics}
\label{Evaluation Metrics}
In the field of visual attention and saliency prediction, many consistency metrics are generally adopted to evaluate the performance of saliency algorithms. We select seven commonly used metrics including AUC-J, sAUC, CC, IG, KL, NSS, and SIM \cite{bylinskii2018different}. These saliency evaluation metrics can be categorized into two types including the location-based metrics and the distribution-based metrics \cite{riche2013saliency_46, bylinskii2018different, kummerer2015information_48}, as summarized in Table \ref{metrics}. Location-based metrics consider saliency values at discrete fixation locations, while distribution-based metrics consider both ground truth fixation density maps and predicted saliency maps as continuous distributions.

\begin{table}[t]
    \centering
    \caption{Different metrics use different formats of ground truth for evaluating saliency models.}
    \vspace{-0.8em}
    \setstretch{1.1}
    \resizebox{0.45\textwidth}{!}{
    \begin{tabular}{ccc}
    \Xhline{1pt}
    Metrics       & Location-based       & Distribution-based \\ \hline
    Similarity    & AUC-J, sAUC, NSS, IG & SIM, CC            \\
    Dissimilarity &    -               & KL                 \\ \Xhline{1pt}
    \end{tabular}
    }
    \label{metrics}
    \vspace{-1em}
\end{table}

\begin{table*}[t]
    \centering
    \caption{Quantitative comparison results of different models, which are trained on the training set of the SALICON database \cite{SALICON_database}, and tested on the validation set of the SALICON database \cite{SALICON_database} as well as the MIT1003 database \cite{MIT1003}, respectively. We \textbf{bold} the best result and \underline{underline} the second-best result, the same rule is applied to all tables below.}
    \vspace{-0.8em}
    \setstretch{1.1}
    \resizebox{0.73\textwidth}{!}{
    \begin{tabular}{l|ccccccc|ccccccc}
    \toprule[1.5pt]
    Training    & \multicolumn{14}{c}{SALICON}   \\ \midrule
    Testing     & \multicolumn{7}{c|}{SALICON}    & \multicolumn{7}{c}{MIT1003}   \\ \midrule
    Model\textbackslash Metric & AUC-J↑ & sAUC↑  & CC↑    & IG↑     & KL↓    & NSS↑   & SIM↑ & AUC-J↑ & sAUC↑  & CC↑    & IG↑     & KL↓    & NSS↑   & SIM↑   \\ \midrule
    SALICON\cite{SALICON}                 & 0.8304  & 0.6569 & 0.7486 & 34.8575 & 5.7630  & 1.5003 & 0.6664 & 0.8380  & 0.6816 & 0.5012 & 33.5560 & 6.4333 & 1.6648 & 0.4290 \\
    ML-Net\cite{ML-Net}                & 0.8094  & 0.5857 & 0.6746 & 34.7837 & 5.7142 & 1.5021 & 0.5858 & 0.8241  & 0.6226 & 0.4836 & 33.5850 & 6.4046 & 1.7424 & 0.3675\\
    SalGAN \cite{SalGAN}                 & 0.8601  & 0.6569 & 0.8601 & \underline{35.2235} & 5.4493 & 1.7989 & 0.7520  & 0.8779  & 0.7051 & 0.6245 & 33.7281 & 6.3003 & 2.1328 & 0.5046\\
    SAM-VGG \cite{SAM}             & 0.8524  & 0.6092 & 0.8247 & 34.8683 & 5.6555 & 1.7637 & 0.7300   & 0.8539  & 0.6539 & 0.5788 & 32.9439 & 6.8819 & 2.0078 & 0.4786\\
    SAM-ResNet  \cite{SAM}           & 0.8543  & 0.6019 & 0.8398 & 34.9310  & 5.6120  & 1.8121 & 0.7376 & 0.8694  & 0.6639 & 0.6153 & 33.6811 & 6.3315 & 2.1320 & 0.4908\\
    GazeGAN\cite{GazeGAN}              & 0.8522  & 0.6175 & 0.8056 & 35.1258 & 5.9386 & 1.6241 & 0.7014 & 0.8619  & 0.7101 & 0.5423 & 33.7958 & 6.6618 & 1.7864 & 0.4498\\
    VQSal\cite{duan2022saliency}  & \underline{0.8627} & 0.6295 & \underline{0.8688} & 35.1084 & 5.4891 & \underline{1.8599} & \underline{0.7653} & \underline{0.8782}  & 0.6888 & \underline{0.6370} & 33.6711 & 6.3331 & \underline{2.2142} & \underline{0.5170}\\
        
    TranSalNet\cite{TranSalNet} 	&0.8496 &0.6228	&0.8325	&35.1132	&5.4986	&1.7996	&0.7365 & 0.8693  & 0.7168 & 0.5520 & \underline{33.8875} & 6.1959 & 1.7829 & 0.4546\\    
    TempSAL\cite{TempSAL}	    &0.8395	&0.6347	&0.8245	&35.0480	&5.5310	&1.7656	&0.7105 & 0.8497  & 0.6714 & 0.5203 & 33.8285 & 6.2373 & 1.7650 & 0.4362\\
    GSGNet \cite{GSGNet}	    &0.8548	&0.6354	&0.8466	&35.2124	&\underline{5.4170}	&1.8339	&0.7444 & 0.8656  & 0.6813 & 0.5795 & 33.8084 & \underline{6.1911} & 1.9846 & 0.4706\\ \hline
        
    ALBEF \cite{ALBEF}      &0.8564	&0.6524	&0.8469	&34.9368	&5.6080	&1.7368	&0.7438 & 0.8743	& 0.7043	& 0.5958	& 33.6031	& 6.3841	& 1.9971	&0.4822\\
    BLIP\cite{BLIP}	    &0.8603	&\underline{0.6578}	&0.8649	&35.0951	&5.4983	&1.8070	&0.7598 & 0.8768  & \underline{0.7181} & 0.6172 & 33.7618 & 6.2746 & 2.0964 & 0.4962\\
    ImageBind\cite{ImageBind}  &0.8558	&0.6552	&0.8423	&35.1582	&5.4546	&1.7248	&0.7328  &0.8728	&0.7171	&0.5925	&33.8663	&6.2089	&1.9851	&0.4656\\
    TGSal                  & \textbf{0.8658}  & \textbf{0.6650}  & \textbf{0.8816} & \textbf{35.2529} & \textbf{5.3767} & \textbf{1.8606} & \textbf{0.7731} & \textbf{0.8873}  & \textbf{0.7230} & \textbf{0.6632} & \textbf{33.8931} & \textbf{6.1815} & \textbf{2.2856} & \textbf{0.5259}\\
    \bottomrule[1.5pt]
    \end{tabular}
    }
    \label{salicon}
    \vspace{-1em}
\end{table*}

\subsubsection{Implementation Details}
We first conduct experiments on SALICON \cite{SALICON_database} to validate the general effectiveness of our TGSal model. We train the model only on the training set of SALICON and test the model performance on the validation set of it. We further conduct a cross-dataset evaluation experiment by training the model on the training set of the SALICON \cite{SALICON_database} and testing it on MIT1003 \cite{MIT1003}. All comparative deep learning models mentioned in the Sections \ref{model_train1} and \ref{model_train2} are trained and tested using the same way. During the training process, we set the text input of our TGSal model as empty and freeze the text feature extraction module and the image feature extraction module. The model is trained on SALICON for a total of 20 epochs, with an initial learning rate of 1$e^{-4}$, and gradually reduces the learning rate to 1$e^{-5}$ with the cosine annealing \cite{loshchilov2016sgdr, duan2023masked, duan2022develop}. The hyperparameters $\alpha$ and $\beta$ in the loss function are empirically set as 0.06 and 1 respectively, to balance the contributions of the two loss items. 

Then we conduct experimental validation on the SJTU-TIS database. As aforementioned, our database contains five types of visual attention data, which includes visual attention on pure images, visual attention on general scenario text-guided images, visual attention on images guided by specified text descriptions of salient objects, visual attention on images guided by specified descriptions of non-salient objects, visual attention on images guided by common descriptions containing both salient and non-salient objects. 
Therefore, it is necessary to consider both the individual training strategy and the joint training strategy on five conditions. Firstly, We use the weights pretrained on SALICON \cite{SALICON_database} as the initial weights, and then individually finetune the model on five conditions. Secondly, we still use the weights pretrained on SALICON \cite{SALICON_database} as the initial weights, but then jointly finetune the model on five conditions.
 Finally, in order to validate that the proposed database can independently support training and evaluation, we further perform the joint training procedure on five categories, however, without pretraining on SALICON \cite{SALICON_database}.
During the experiments, we randomly split each group of data into two sets: a training set (80\% of data) and a testing set (20\% of data), without overlapping during the experiment.
Each experiment is repeated 10 times to prevent performance bias in each group, and mean results are recorded for comparison.

\vspace{-1em}

\subsection{Comparison with State-of-the-art Methods on Pure Image Saliency Databases}
\label{model_train1}
We first compare the performance of the proposed TGSal with ten state-of-the-art saliency models including SALICON \cite{SALICON}, ML-Net \cite{ML-Net}, SalGAN \cite{SalGAN}, SAM-VGG \cite{SAM}, SAM-ResNet \cite{SAM}, GazeGAN \cite{GazeGAN}, VQSal \cite{duan2022saliency}, TranSalNet \cite{TranSalNet}, TempSAL \cite{TempSAL}, GSGNet \cite{GSGNet} and three text-image pretraining models including ALBEF \cite{ALBEF}, BLIP \cite{BLIP}, and ImageBind \cite{ImageBind} on the SALICON database. For fair comparison, all these comparative models are retrained on SALICON using their officially released code. Table \ref{salicon} shows the performance of the baseline models and our proposed model on the SALICON database. It can be observed that our TGSal model achieves the best performance compared to other state-of-the-art models under the pure image condition, which manifests the superiority of the proposed model.

In order to demonstrate the generalization ability of the proposed model for unseen images, we further conduct the cross-dataset evaluation experiment by training the model on the training set of the SALICON \cite{SALICON_database} and testing it on MIT1003 \cite{MIT1003}. As shown in Table \ref{salicon}, our proposed TGSal model still performs better than other state-of-the-art models on this cross-dataset validation experiment, which further manifests the superiority of our model.

\begin{figure*}[!t]
\centering
\includegraphics[width=6in]{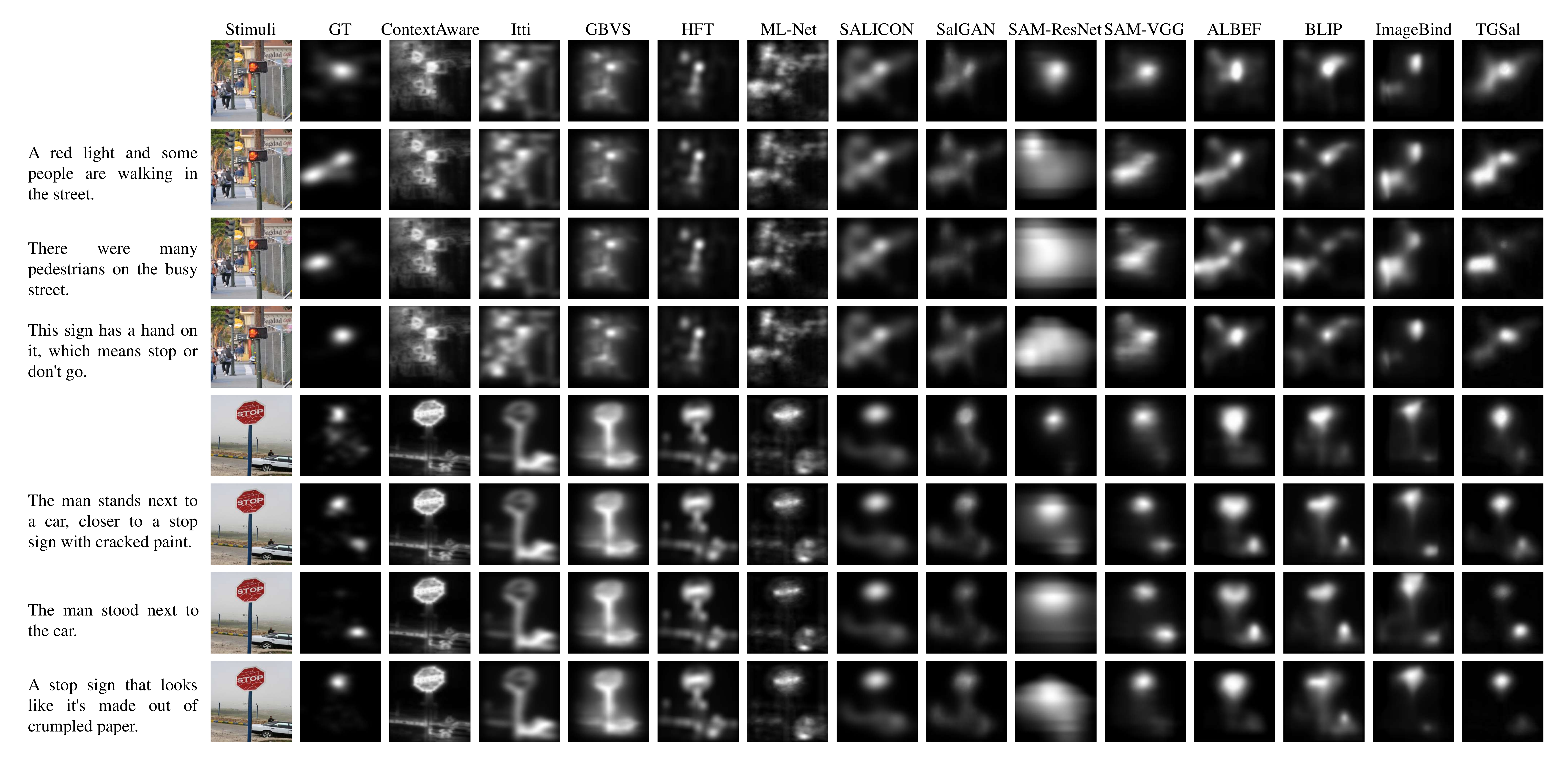}
\vspace{-1em}
\caption{Qualitative comparison results for state-of-the-art saliency prediction models on our SJTU-TIS database.}
\label{output}
\vspace{-1em}
\end{figure*}

\begin{table*}[!t]
    \centering
    \caption{Quantitative comparisons between our proposed TGSal model and benchmark methods under pure image condition and the general scenario description condition. All models are first pretrained on SALICON \cite{SALICON_database}, and then finetuned individually on different conditions of SJTU-TIS.}
    \vspace{-0.8em}
    \setstretch{1.1}
    \resizebox{0.73\textwidth}{!}{
        \begin{tabular}{l|ccccccc|ccccccc}
    \toprule[1.5pt]
    Type                        & \multicolumn{7}{c|}{Pure images}                                       & \multicolumn{7}{c}{General scenario descriptions}                                       \\ \midrule
    Model\textbackslash{}Metric & AUC-J↑ & sAUC↑  & CC↑    & IG↑     & KL↓     & NSS↑   & SIM↑   & AUC-J↑ & sAUC↑  & CC↑    & IG↑     & KL↓     & NSS↑   & SIM↑   \\ \midrule
    
    IT \cite{IT}                       & 0.7713  & 0.5913 & 0.4931 & 34.6916 & 6.8440   & 1.0248 & 0.5329 & 0.7900    & 0.6052 & 0.4950  & 34.7876 & 6.7735  & 1.1235 & 0.5045 \\
    AIM \cite{AIM}                         & 0.7119  & 0.6145 & 0.3494 & 34.3621 & 7.0724  & 0.7551 & 0.4635 & 0.7318  & 0.6285 & 0.3594 & 34.4300   & 7.0213  & 0.8376 & 0.4317 \\
    GBVS \cite{GBVS}                        & 0.8188  & 0.6169 & 0.6255 & 34.9341 & 6.6759  & 1.2715 & 0.5942 & 0.8327  & 0.6249 & 0.6140  & 35.0264 & 6.6080   & 1.3578 & 0.5621 \\
    SMVJ \cite{SMVJ}                        & 0.6497  & 0.5434 & 0.2412 & 34.1852 & 7.1880   & 0.5610  & 0.4384 & 0.6522  & 0.5440  & 0.2277 & 34.2292 & 7.1605  & 0.5806 & 0.4033 \\
    SUN \cite{SUN}                         & 0.6453  & 0.5465 & 0.2632 & 34.2230 & 7.0723   & 0.5728  & 0.4582 & 0.6628  & 0.5527  & 0.2923 & 34.2836 & 7.0273  & 0.5927 & 0.4263 \\
    Hou \cite{Hou_NIPS}                   & 0.6085  & 0.5017 & 0.1230  & 20.5622 & 16.6378 & 0.3253 & 0.2556 & 0.6110   & 0.5016 & 0.1196 & 20.9027 & 16.3977 & 0.3417 & 0.2474 \\
    SeR \cite{SeR}                         & 0.6246  & 0.5323 & 0.1905 & 33.8000    & 7.4620   & 0.4642 & 0.4093 & 0.6264  & 0.5314 & 0.1809 & 33.8824 & 7.4009  & 0.4816 & 0.3807 \\
    CA \cite{CA}                & 0.7462  & 0.5974 & 0.4290  & 34.5718 & 6.9270   & 0.9349 & 0.5085 & 0.7602  & 0.6045 & 0.4290  & 34.6780  & 6.8498  & 1.0194 & 0.4812 \\
    HFT \cite{HFT}                         & 0.8058  & 0.5834 & 0.5997 & 34.9208 & 6.6851  & 1.2851 & 0.5858 & 0.8248  & 0.5945 & 0.6026 & 35.0600   & 6.5847  & 1.4177 & 0.5630  \\
    CovSal \cite{CovSal}                      & 0.8288  & 0.5629 & 0.6557 & 34.3128 & 7.1066  & 1.4102 & 0.6112 & 0.8448  & 0.5672 & 0.6500   & 34.7237 & 6.8178  & 1.5139 & 0.6046 \\    \midrule
    SALICON\cite{SALICON}                   & 0.8397  & \underline{0.6473} & 0.7221 & 35.0371 & 6.6070   & 1.5644 & 0.6578 & 0.8574  & 0.6751 & 0.6817 & 35.3153 & 6.4057  & 1.6872 & 0.6014 \\
    ML-Net \cite{ML-Net}                      & 0.8070   & 0.6056 & 0.6139 & 34.8630  & 6.7277  & 1.4676 & 0.5844 & 0.8332  & 0.6158 & 0.6086 & 35.0254 & 6.6066  & 1.6286 & 0.5359 \\
    SalGAN \cite{SalGAN}                   & 0.8625  & 0.6433 & 0.8397 & 35.1372 & 6.5376  & 1.8514 & 0.7461 & 0.8838  & 0.7029 & 0.7703 & 35.4254 & 6.3294  & 1.8959 & 0.6744 \\
    SAM-VGG \cite{SAM}                    & \underline{0.8644}  & 0.6290  & \underline{0.8635} & \underline{35.2599} & 6.5456  & 1.8835 & 0.7403 & 0.8892  & 0.6706 & 0.8321 & 35.5403 & 6.2497  & 2.0446 & 0.6954 \\
    SAM-ResNet \cite{SAM}                 & 0.8601  & 0.6393 & 0.8469 & 34.6907 & 6.8471  & \underline{1.8883} & 0.7331 & 0.8839  & 0.6512 & 0.8122 & 35.1381 & 6.5285  & 2.0484 & 0.6863 \\
    GazeGAN \cite{GazeGAN}                   & 0.8381  & 0.6447 & 0.7521 & 35.1043 & 6.9962  & 1.4737 & 0.6856 & 0.8642  & 0.6956 & 0.7160  & 35.2426 & 7.1267  & 1.5576 & 0.6192 \\
    VQSal\cite{duan2022saliency}  & 0.8624 & 0.6404 & 0.8531 & 34.8010 & 6.7707 & 1.8606 & \underline{0.7531} & 0.8884 & 0.6754 & 0.8215 & 35.5494 & 6.3226 & 2.0231 & 0.6919 \\
    TranSalNet\cite{TranSalNet}  	&0.8576	&0.6461	&0.7641	&35.2313	&6.4731	&1.7185	&0.6833 &0.8738	&0.6852	&0.7063	&35.4594	&6.3079	&1.7939	&0.6299\\
    TempSAL\cite{TempSAL} 	    &0.8463	&0.6311	&0.7938	&35.2208	&6.4797	&1.6241	&0.7147 &0.8768	&0.6587	&0.7812	&35.5270	&6.2589	&1.8374	&0.6868\\
    GSGNet\cite{GSGNet}  	    &0.8573	&0.6419	&0.8385	&35.2436	&6.4945	&1.7591	&0.7485 &0.8890	&0.6847	&0.8281	&35.4826	&6.2510	&2.0018	&0.6898\\    
    TGSal-I               & \textbf{0.8672}  & \textbf{0.6551} & \textbf{0.8736} & \textbf{35.2985} & \textbf{6.4258}  & \textbf{1.9093} & \textbf{0.7711} & \underline{0.8924}  & \underline{0.7034} & \underline{0.8370} & 35.5542 & 6.2301  & \underline{2.0552} & 0.7011 \\ \midrule
    
    ALBEF\cite{ALBEF}  &0.8577	&0.6440	&0.8361	&35.1847	&6.5047	&1.7731	&0.7403 &0.8876	&0.7025	&0.8253	&35.5008	&6.2078	&2.0102	&0.7083 \\        
    BLIP\cite{BLIP}  &0.8573	&0.6414	&0.8326	&35.2552	&\underline{6.4558}	&1.7946	&0.7391 &0.8877	&0.6871	&0.8195	&35.5481	&\underline{6.2050}	&2.0327	&\underline{0.7124} \\
    ImageBind\cite{ImageBind} 	&0.8512	&0.6371	&0.8360	&34.7934	&6.7759	&1.8015	&0.7361 &0.8856	&0.6945	&0.8201	&\underline{35.5830}	&6.2201	&1.9835	&0.6997 \\
    TGSal                 &\textbf{0.8672}  & \textbf{0.6551} & \textbf{0.8736} & \textbf{35.2985} & \textbf{6.4258}  & \textbf{1.9093} & \textbf{0.7711} & \textbf{0.8946}  & \textbf{0.7057} & \textbf{0.8501} & \textbf{35.6382} & \textbf{6.1819}  & \textbf{2.0916} & \textbf{0.7166} \\ 
    \bottomrule[1.5pt]    
    \end{tabular}
    }
    \label{pure and whole}
    \vspace{-1em}
\end{table*}

\begin{table*}[!t]
\vspace{-0.5em}
    \centering
    \caption{Quantitative comparisons between our proposed TGSal model and benchmark methods under the conditions of specified descriptions of salient objects and specified descriptions of non-salient objects. All models are first pretrained on SALICON \cite{SALICON_database}, and then finetuned individually on different conditions of SJTU-TIS.}
    \vspace{-0.8em}
    \setstretch{1.1}
    \resizebox{0.73\textwidth}{!}{
        \begin{tabular}{l|ccccccc|ccccccc}
        \toprule[1.5pt]
        Type                        & \multicolumn{7}{c|}{Specified descriptions of salient objects}                                    & \multicolumn{7}{c}{Specified descriptions of non-salient objects}                                 \\ \midrule
        Model\textbackslash{}Metric & AUC-J↑ & sAUC↑  & CC↑    & IG↑     & KL↓     & NSS↑   & SIM↑   & AUC-J↑ & sAUC↑  & CC↑    & IG↑     & KL↓     & NSS↑   & SIM↑   \\ \midrule
        IT \cite{IT}                        & 0.7790  & 0.6026 & 0.4293 & 34.7198 & 6.8789  & 1.0672 & 0.4641 & 0.7625  & 0.5840 & 0.3552 & 34.6425 & 6.9461  & 0.9756 & 0.4370 \\
        AIM \cite{AIM}                         & 0.7192  & 0.6203 & 0.3054 & 34.3793 & 7.1149  & 0.7771 & 0.4045 & 0.7347  & 0.6331 & 0.2999 & 34.4185 & 7.1014  & 0.8349 & 0.3982 \\
        GBVS \cite{GBVS}                        & 0.8173  & 0.6160 & 0.5149 & 34.9186 & 6.7411  & 1.2566 & 0.5064 & 0.7637  & 0.5631 & 0.3754 & 34.6535 & 6.9384  & 0.9815 & 0.4560 \\
        SMVJ \cite{SMVJ}                        & 0.6560  & 0.5445 & 0.2106 & 34.2260 & 7.2182  & 0.5809 & 0.3827 & 0.6665  & 0.5501 & 0.2050 & 34.2559 & 7.2148  & 0.6178 & 0.3748 \\
        SUN \cite{SUN}                         & 0.6660  & 0.5566 & 0.2283 & 34.2843 & 7.1723  & 0.5972 & 0.3957 & 0.6723  & 0.5632 & 0.2273 & 34.3184 & 7.1836  & 0.6365 & 0.3827 \\
        Hou \cite{Hou_NIPS}                   & 0.6120  & 0.5017 & 0.1073 & 20.4085 & 16.7987 & 0.3283 & 0.2301 & 0.6217  & 0.5023 & 0.1027 & 21.0527 & 16.3658 & 0.3666 & 0.2243 \\
        SeR \cite{SeR}                         & 0.6322  & 0.5354 & 0.1695 & 33.8826 & 7.4592  & 0.4852 & 0.3606 & 0.6461  & 0.5418 & 0.1720 & 33.9586 & 7.4201  & 0.5366 & 0.3540 \\
        CA \cite{CA}                & 0.7477  & 0.6007 & 0.3604 & 34.5780 & 6.9769  & 0.9322 & 0.4397 & 0.7494  & 0.5992 & 0.3335 & 34.5879 & 6.9841  & 0.9433 & 0.4269 \\
        HFT \cite{HFT}                         & 0.8063  & 0.5824 & 0.4962 & 34.9212 & 6.7393  & 1.2786 & 0.5014 & 0.7688  & 0.5561 & 0.3637 & 34.6708 & 6.9265  & 1.0101 & 0.4483 \\
        CovSal \cite{CovSal}                      & 0.8183  & 0.5477 & 0.4845 & 34.2583 & 7.1988  & 1.2429 & 0.5106 & 0.7414  & 0.5198 & 0.2870 & 32.5899 & 8.3688  & 0.7617 & 0.4072 \\ \midrule
        
        SALICON \cite{SALICON}                     & 0.8361  & 0.6487 & 0.5452 & 35.0417 & 6.6516  & 1.5154 & 0.5164 & 0.8133  & 0.6266 & 0.4438 & 34.8150 & 6.8100  & 1.2487 & 0.4799 \\
        ML-Net \cite{ML-Net}                     & 0.8217  & 0.5921 & 0.4874 & 34.9345 & 6.7259  & 1.5190 & 0.4716 & 0.8122  & 0.5822 & 0.4273 & 34.8340 & 6.7969  & 1.4189 & 0.4532 \\
        SalGAN \cite{SalGAN}                      & \underline{0.8713}  & 0.6493 & 0.6346 & 35.2968 & 6.4748  & 1.7668 & 0.5764 & 0.8347  & 0.6285 & 0.5240 & 34.9519 & 6.7152  & 1.5354 & 0.5236 \\
        SAM-VGG \cite{SAM}                    & 0.8500  & 0.6229 & 0.6123 & 34.7479 & 6.8553  & 1.7561 & 0.5808 & 0.8378  & 0.6345 & 0.5327 & 34.9263 & 6.7329  & 1.6626 & 0.5420 \\
        SAM-ResNet \cite{SAM}                 & 0.8182  & 0.6386 & 0.5454 & 33.8274 & 7.4933  & 1.3596 & 0.5428 & 0.7873  & 0.6139 & 0.4480 & 34.0277 & 7.3558  & 1.1538 & 0.4905 \\
        GazeGAN \cite{GazeGAN}                     & 0.8109  & 0.6374 & 0.4986 & 34.8390 & 7.3259  & 1.2032 & 0.4998 & 0.7368  & 0.5728 & 0.3344 & 34.3832 & 7.2974  & 0.8382 & 0.4468 \\
        VQSal \cite{duan2022saliency} & 0.8603 & 0.6258 & 0.6302 & 35.1000 & 6.6112 & 1.7917 & 0.5878 & 0.8262 & 0.6099 & 0.5238 & 34.3067 & 7.1624 & 1.6233 & 0.5417 \\
        TranSalNet\cite{TranSalNet} 	&0.8636	&0.6417	&0.6086	&35.2680	&6.5094	&1.7354	&0.5497 &0.8308	&0.6360	&0.5033	&35.0546	&6.6594	&1.5248	&0.5073 \\
        TempSAL\cite{TempSAL}	    &0.8156	&0.6095	&0.5409	&34.9678	&6.7028	&1.4287	&0.5435 &0.7693	&0.6031	&0.3865	&34.6256	&6.9414	&1.0188	&0.4654 \\
        GSGNet\cite{GSGNet}  	&0.8443	&0.6304	&0.6045	&35.2540	&6.5045	&1.7125	&0.5839 &0.8238	&0.6126	&0.4890	&35.0044	&6.6788	&1.4432	&0.5118 \\ 
        TGSal-I               & 0.8696  & 0.6509 & 0.6420 & 35.3245 & 6.4249  & 1.9067 & \underline{0.5952} & 0.8459  & \underline{0.6376} & 0.5609 & 35.2360 & 6.5183  & 1.7116 & 0.5440 \\ \midrule
        ALBEF\cite{ALBEF} &0.8660	&0.6426	&0.6446	&34.8139	&6.8095	&1.9194	&0.5827 &0.8437	&0.6349	&0.5565	&35.0303	&6.6608	&1.6807	&0.5382 \\    
        BLIP\cite{BLIP}        &0.8680	&0.6485	&\underline{0.6513}	&\underline{35.4132}	&\underline{6.3941}	&\underline{1.9524}	&0.5821  &\underline{0.8593}	&0.6341	&\underline{0.5821}	&\underline{35.2579}	&\underline{6.5030}	&\underline{1.7581}	&\underline{0.5451} \\ 
        ImageBind\cite{ImageBind}	&0.8501	&\underline{0.6567}	&0.6362	&35.0579	&6.6404	&1.8073	&0.5827  &0.8238	&0.6286	&0.5192	&34.9103	&6.7440	&1.4793	&0.5212 \\    
        TGSal                 & \textbf{0.8728}  & \textbf{0.6790} & \textbf{0.6702} & \textbf{35.4654} & \textbf{6.3579}  & \textbf{1.9615} & \textbf{0.6023} & \textbf{0.8601}  & \textbf{0.6405} & \textbf{0.6210} & \textbf{35.3746} & \textbf{6.4222}  & \textbf{1.9288} & \textbf{0.5774} \\ 
        \bottomrule[1.5pt]
        \end{tabular}
    }
    \label{salient and non-salient}
    \vspace{-1em}
\end{table*}

\begin{table*}[!t]
    \centering
    \caption{Quantitative comparisons between our proposed TGSal model and benchmark methods under the condition of common descriptions containing both salient and non-salient objects and averaged among all conditions. All models are first pretrained on SALICON \cite{SALICON_database}, and then finetuned individually on different conditions of SJTU-TIS.}
    \vspace{-0.8em}
    \setstretch{1.1}
    \resizebox{0.73\textwidth}{!}{
    \begin{tabular}{l|ccccccc|ccccccc}
        \toprule[1.5pt]
        Type    & \multicolumn{7}{c|}{Common descriptions contain both salient and non-salient objects }   & \multicolumn{7}{c}{Averaged among all conditions}     \\ \midrule
        Model\textbackslash{}Metirc & AUC-J↑ & sAUC↑  & CC↑    & IG↑     & KL↓     & NSS↑   & SIM↑ & AUC-J↑ & sAUC↑  & CC↑    & IG↑     & KL↓     & NSS↑   & SIM↑   \\ \midrule
        IT \cite{IT}                        & 0.7771  & 0.5963 & 0.4671 & 34.7056 & 6.8695  & 1.0487 & 0.5092 &0.7752	&0.5951	&0.4555	&34.7065	&6.8593	&1.0441	&0.4968\\
        AIM \cite{AIM}                         & 0.7346  & 0.6324 & 0.3629 & 34.4159 & 7.0703  & 0.8344 & 0.4488 &0.7240	&0.6239	&0.3377	&34.3947	&7.0755	&0.7990	&0.4350\\
        GBVS \cite{GBVS}                       & 0.7965  & 0.5962 & 0.5311 & 34.8138 & 6.7945  & 1.1479 & 0.5435 &0.8080	&0.6057	&0.5477	&34.8801	&6.7390	&1.2145	&0.5427\\
        SMVJ \cite{SMVJ}                        & 0.6668  & 0.5507 & 0.2483 & 34.2548 & 7.1813  & 0.6197 & 0.4227 &0.6568	&0.5460	&0.2290	&34.2227	&7.1918	&0.5868	&0.4101\\
        SUN \cite{SUN}                         & 0.6823  & 0.5728 & 0.2823 & 34.3583 & 7.0827  & 0.6378 & 0.4426 &0.6623	&0.5564	&0.2594	&34.2818	&7.1018	&0.6016	&0.4273\\
        Hou \cite{Hou_NIPS}                   & 0.6192  & 0.5020  & 0.1239 & 20.8938 & 16.4431 & 0.3607 &0.2504 &0.6135	&0.5018	&0.1166	&20.7304	&16.5468&0.3413 &0.2439\\
        SeR \cite{SeR}                         & 0.6459  & 0.5426 & 0.2061 & 33.9627 & 7.3844  & 0.5414 &0.4006 &0.6333	&0.5360	&0.1849	&33.8811	&7.4314	&0.4955	&0.3858\\
        CA \cite{CA}                & 0.7557  & 0.6068 & 0.4185 & 34.6170  & 6.9311  & 0.9794 &0.4898 &0.7509	&0.6010	&0.3999	&34.6008	&6.9327	&0.9574	&0.4758\\
        HFT \cite{HFT}                         & 0.7954  & 0.5751 & 0.5201 & 34.8369 & 6.7785  & 1.1890  &0.5403 &0.8012	&0.5792	&0.5303	&34.8884	&6.7332	&1.2443	&0.5374\\
        CovSal \cite{CovSal}                      & 0.7849  & 0.5363 & 0.4626 & 33.5774 & 7.6515  & 1.0324 &0.5087 &0.8078	&0.5495	&0.5326	&33.9625	&7.3750	&1.2286	&0.5423\\ \midrule        
        SALICON\cite{SALICON}                     & 0.8347  & 0.6560  & 0.6123 & 35.0291 & 6.6435  & 1.4733 & 0.5852 
                                &0.8368	&0.6502	&0.6212	&35.0459	&6.6208	&1.5089	&0.5831\\
        ML-Net\cite{ML-Net}                      & 0.8241  & 0.6024 & 0.5552 & 34.9594 & 6.6917  & 1.5216 & 0.5310  
                                    &0.8175	&0.6006	&0.5511	&34.9132	&6.7128	&1.5039	&0.5268\\
        SalGAN\cite{SalGAN}                     & 0.8639  & 0.6435 & 0.7002 & 35.1519 & 6.5583  & 1.7168 & 0.6438 
                                    &0.8631	&0.6518	&0.7181	&35.1834	&6.5255	&1.7696	&0.6517\\
        SAM-VGG \cite{SAM}                    & 0.8488  & 0.6261 & 0.6865 & 34.7915 & 6.8082  & 1.7396 & 0.6455
                                    &0.8591	&0.6354	&0.7318	&35.0876	&6.6229	&1.8283	&0.6574\\
        SAM-ResNet\cite{SAM}                & 0.8064  & 0.6100   & 0.5901 & 34.0400   & 7.3290   & 1.3395 & 0.5848 
                                    &0.8360	&0.6321	&0.6816	&34.4024	&7.0668	&1.6130	&0.6284\\
        GazeGAN \cite{GazeGAN}                    & 0.7989  & 0.6251 & 0.5437 & 34.7767 & 7.1847  & 1.1823 & 0.5663 
                                    &0.8145	&0.6367	&0.5995	&34.9084	&7.1545	&1.2881	&0.5839\\
        VQSal   \cite{duan2022saliency}                     & 0.8629 & 0.6344 & 0.7041 & 35.1222 & 6.5789 & 1.8296 & 0.6588 
                                    &0.8604	&0.6377	&0.7310	&34.9467	&6.7028	&1.8315	&0.6644\\
        TranSalNet 	\cite{TranSalNet}                &0.8462	&0.6534	&0.6250	&35.1361	&6.5741	&1.5924	&0.5912 
                                    &0.8549	&0.6514	&0.6619	&35.2301	&6.4995	&1.6806	&0.6075\\
        TempSAL	\cite{TempSAL}                    &0.8067	&0.6245	&0.5748	&34.8571	&6.7626	&1.2545	&0.5715 
                                    &0.8268	&0.6263	&0.6452	&35.0699	&6.6042	&1.4646	&0.6161\\
        GSGNet \cite{GSGNet}	                    &0.8462	&0.6210	&0.6583	&35.1198	&6.5380	&1.6262	&0.5895 
                                    &0.8530	&0.6388	&0.7095	&35.2247	&6.4936	&1.7170	&0.6453\\
        TGSal-I                     & 0.8652  & 0.6601 & 0.7118 & \underline{35.1726} & 6.5283  & \underline{1.8323} & \underline{0.6618} 
                                    &\underline{0.8679}	&\underline{0.6604}	&\underline{0.7498}	&35.3141	&6.4255	&\underline{1.8874}	&\underline{0.6741} \\ \midrule
        
        ALBEF  \cite{ALBEF}                  &0.8625	&0.6604	&0.7244	&35.1332	&6.5713	&1.7957	&0.6530 
                                &0.8625	&0.6547	&0.7372	&35.1413	&6.5431	&1.8254	&0.6605\\
        BLIP\cite{BLIP}	                      &\underline{0.8672}	&\underline{0.6612}	&\underline{0.7284}	&35.1673	&\underline{6.5090}	&1.8002	&0.6512	
                                    &0.8661	&0.6523	&0.7411	&\underline{35.3162}	&\underline{6.4205}	&1.8554	&0.6615 \\
        ImageBind   \cite{ImageBind}             &0.8480	&0.6550	&0.6890	&35.1466	&6.5620	&1.6595	&0.6355  &0.8517	&0.6515	
                                &0.7228	&35.0474	&6.6197	&1.7554	&0.6519 \\
        TGSal           & \textbf{0.8684}  & \textbf{0.6793}  & \textbf{0.7317}  & \textbf{35.3196}  & \textbf{6.4421}   & \textbf{1.8449}  & \textbf{0.6696} & \textbf{0.8717}	&\textbf{0.6691}	&\textbf{0.7700}	&\textbf{35.3991}	&\textbf{6.3760}	&\textbf{1.9409}	&\textbf{0.6847} \\         
        \bottomrule[1.5pt]
        \end{tabular}
    }
    \label{common}
    \vspace{-1em}
\end{table*}

\begin{table}[!t]

    \centering
    \caption{Comparison of performance improvement between the TGSal and the TGSal-I model under four conditions, respectively.}
    \vspace{-0.8em}
    \setstretch{1.1}
    \setlength{\tabcolsep}{1mm}
    \resizebox{0.43\textwidth}{!}{
    \begin{tabular}{c|ccccccc}
    \toprule[1.5pt]
    Type\textbackslash{}Metric & AUC-J↑ & sAUC↑  & CC↑    & IG↑     & KL↓    & NSS↑   & SIM↑   \\ \midrule
    Type 1                      & 0.247\%  & 0.327\% & 1.565\%  & 0.236\% & 0.774\% & 1.771\%  & 2.211\% \\ 
    Type 2                      & 0.368\%  & \textbf{4.317\%} & 4.393\%  & 0.399\% & 1.043\% & 2.847\%  & 1.193\% \\ 
    Type 3                      & \textbf{1.679\%}  & 0.455\% & \textbf{10.715\%} & 0.393\% & \textbf{1.474\%} & \textbf{12.690\%} & \textbf{6.140\%} \\ 
    Type 4                      & 0.370\%  & 2.909\% & 2.796\%  & \textbf{0.418\%} & 1.320\% & 0.688\%  & 1.179\% \\ 
    \bottomrule[1.5pt]
    \end{tabular}
    }
    \label{grow}
    \vspace{-2em}
\end{table}

\begin{table*}[!t]
\vspace{-0.5em}
    
    \centering
    \caption{Quantitative comparisons between our proposed TGSal model and benchmark methods under the pure image condition and the general scenario description condition. All models are first pretrained on SALICON \cite{SALICON_database}, and then jointly finetuned on all conditions of SJTU-TIS.}
    \label{SALICON_jointly_pure_whole}
    \vspace{-0.8em}
    \setstretch{1.1}
    \resizebox{0.73\textwidth}{!}{
    \begin{tabular}{l|ccccccc|ccccccc}
    \toprule[1.5pt]
    Type                        & \multicolumn{7}{c|}{Pure images}                                       & \multicolumn{7}{c}{General scenario descriptions}                                       \\ \midrule
    Model\textbackslash{}Metirc & AUC-J↑ & sAUC↑  & CC↑    & IG↑     & KL↓     & NSS↑   & SIM↑   & AUC-J↑ & sAUC↑  & CC↑    & IG↑     & KL↓     & NSS↑   & SIM↑\\ \midrule
    SALICON\cite{SALICON}     & 0.8223	& 0.6372	& 0.7172	& 34.7283	& 6.6723	& 1.5023	& 0.6382 & 0.8523	& 0.6023	& 0.6532	& 35.2374	& 6.4134	& 1.5723	& 0.5732 \\  
    ML-Net\cite{ML-Net}      & 0.7871 & 0.5727 & 0.5419 & 34.7454 & 6.8092 & 1.3150  & 0.5542 & 0.8181 & 0.5839 & 0.5567 & 34.9876 & 6.6328 & 1.5187 & 0.5118 \\
    SalGAN\cite{SalGAN}      & 0.8551 & 0.6705 & 0.8321 & \underline{35.2596} & 6.4528 & 1.7808 & 0.7303 & 0.8877 & \underline{0.6948} & 0.7911 & 35.5216 & 6.2627 & 1.9243 & 0.6752 \\
    SAM-VGG\cite{SAM}    & 0.8557 & 0.6427 & 0.8230  & 35.0659 & 6.5870  & 1.7796 & 0.7367 & 0.8867 & 0.6581 & 0.8114 & 35.5084 & 6.2718 & 2.0021 & 0.7125 \\
    SAM-ResNet\cite{SAM} & 0.8605 & 0.6315 & 0.8306 & 35.0824 & 6.5758 & \underline{1.8489} & 0.7402 & 0.8898 & 0.6398 & 0.8079 & 35.6015 & 6.2073 & \underline{2.0448} & 0.7158 \\
    GazeGAN\cite{GazeGAN}     & 0.8303 & 0.6653 & 0.7080  & 35.0377 & 6.9076 & 1.3849 & 0.6649 & 0.8709 & 0.6936 & 0.7299 & 35.3843 & 6.7941 & 1.6402 & 0.6465 \\
    VQSal\cite{duan2022saliency}       & \underline{0.8622} & 0.6526 & \underline{0.8433} & 34.8630  & 6.7277 & 1.8278 & \underline{0.7471} & \underline{0.8902} & 0.6663 & 0.8215 & 35.5169 & 6.2659 & 2.0178 & \underline{0.7270}  \\
    TranSalNet\cite{TranSalNet}  & 0.8559 & 0.6656 & 0.7464 & 35.1311 & 6.4725 & 1.7271 & 0.6698 & 0.8744 & 0.6706 & 0.7012 & 35.3926 & 6.3528 & 1.7946 & 0.6135 \\
    TempSAL\cite{TempSAL}     & 0.8436 & 0.6545 & 0.7814 & 35.1829 & 6.5060  & 1.5840  & 0.7054 & 0.8809 & 0.6891 & 0.7903 & 35.5173 & 6.2656 & 1.8425 & 0.6798 \\
    GSGNet\cite{GSGNet}      & 0.8579 & 0.6519 & 0.8277 & 35.1434 & \underline{6.3947} & 1.7619 & 0.7396 & 0.8862 & 0.6731 & 0.8058 & \underline{35.6386} & \underline{6.1816} & 1.9547 & 0.7069 \\ \midrule
    ALBEF-CAM\cite{ALBEF}  & -  & - & - & - & -  & - & - & 0.8040 	&0.5482 	&0.5237 	&34.5185 	&6.9580 	&1.4232 	&0.5019 \\
    BLIP-CAM\cite{BLIP}  & -  & - & - & - & -  & - & - & 0.7786 	&0.5450 	&0.4969 	&34.6611 	&6.8591 	&1.2665 	&0.4986  \\
    ALBEF\cite{ALBEF} &0.8547	&\underline{0.6749}	&0.8132	&34.8568	&6.7320	&1.7090	&0.7236 &0.8885	&0.6845	&0.8142	&35.5104	&6.2705	&1.9704	&0.7047 \\
    BLIP\cite{BLIP} &0.8505	&0.6720	&0.7922	&34.0766	&7.2728	&1.7368	&0.6971	&0.8829	&0.6853	&0.7896	&35.0741	&6.5728	&1.9520	&0.6881 \\
    ImageBind\cite{ImageBind}  &0.8535	&0.6530	&0.8216	&35.2044	&6.4494	&1.7403	&0.7262 &0.8888	&0.6803	&\underline{0.8330}	&35.6168	&6.1967	&2.0124	&0.7041 \\
    TGSal       & \textbf{0.8636} & \textbf{0.6784} & \textbf{0.8671} & \textbf{35.2752} & \textbf{6.3499} & \textbf{1.8651} & \textbf{0.7550}  & \textbf{0.8939} & \textbf{0.6976} & \textbf{0.8443} & \textbf{35.6440}  & \textbf{6.1779} & \textbf{2.0855} & \textbf{0.7345} \\
    \bottomrule[1.5pt]
    \end{tabular}
    }
    \vspace{-1em}
\end{table*}

\begin{table*}[!t]
    
    \centering
    \caption{Quantitative comparisons between our proposed TGSal model and benchmark methods under the conditions of specified descriptions of salient objects and specified descriptions of non-salient objects. All models are first pretrained on SALICON \cite{SALICON_database}, and then jointly finetuned on all conditions of SJTU-TIS.}
    \label{SALICON_jointly_salicon_nonsalicon}
    \vspace{-0.8em}
    \setstretch{1.1}
    \resizebox{0.73\textwidth}{!}{
    \begin{tabular}{l|ccccccc|ccccccc}
    \toprule[1.5pt]
    Type                        & \multicolumn{7}{c|}{Specified descriptions of salient objects}                                    & \multicolumn{7}{c}{Specified descriptions of non-salient objects}                                 \\ \midrule
    Model\textbackslash{}Metirc & AUC-J↑ & sAUC↑  & CC↑    & IG↑     & KL↓     & NSS↑   & SIM↑   & AUC-J↑ & sAUC↑  & CC↑    & IG↑     & KL↓     & NSS↑   & SIM↑\\ \midrule
    SALICON\cite{SALICON}	    &0.8323	&0.6456	&0.5386	&34.8672	&6.7438	&1.4723	&0.5023  &0.8023	&0.6023	&0.4184	&34.6923	&7.0213	&1.1846	&0.4623 \\
    ML-Net\cite{ML-Net}	    &0.8034	&0.5787	&0.4423	&34.8034	&6.8168	&1.3632	&0.4521 & 0.8057
    & 0.5865    & 0.4202    & 34.7900      & 6.8274    & 1.3872     & 0.4447    \\
    SalGAN\cite{SalGAN}	    &0.8697	&0.6774	&0.6435	&35.1939	&6.4968	&1.7882	&0.5763 & 0.8241 & 0.6124    & 0.4742    & \underline{34.8845} & \underline{6.7619} & 1.3416     & 0.5018    \\
    SAM-VGG\cite{SAM}	    &0.8383	&0.6292	&0.5834	&34.9272	&6.7310	&1.6230	&0.5599 & 0.7900      & 0.5770     & 0.4176    & 34.2445    & 7.2055    & 1.1928     & 0.4764    \\
    SAM-ResNet\cite{SAM}	&0.8589	&0.6254	&0.6265	&35.1542	&6.5739	&1.7971	&0.5892 & 0.8203    & 0.5836    & 0.4756    & 34.5574    & 6.9879    & 1.4104     & \underline{0.5087} \\
    GazeGAN\cite{GazeGAN}	    &0.8182	&0.6492	&0.5203	&34.9127	&7.0013	&1.2997	&0.5269 & 0.7486    & 0.5685    & 0.3446    & 34.3859    & 7.1480     & 0.8866     & 0.4528    \\
    VQSal\cite{duan2022saliency}	    &0.8678	&0.6671	&0.6563	&35.1922	&6.5473	&\underline{1.8078}	&0.6064 & 0.8163    & 0.6011    & 0.4759    & 34.1307    & 7.2844    & 1.3878     & 0.5044    \\
    TranSalNet\cite{TranSalNet} 	&0.8660	&0.6748	&0.6033	&35.2414	&6.5149	&1.7836	&0.5340 & 0.8208    & 0.6075    & \underline{0.5009} & 34.7753    & 6.9311    & 1.4692  & 0.4966    \\
    TempSAL\cite{TempSAL}	    &0.8270	&0.6191	&0.5506	&35.0574	&6.6408	&1.4150	&0.5449 &0.7774	&0.5579 & 0.3773    & 34.6301    & 6.9382    & 0.9951     & 0.4712    \\
    GSGNet\cite{GSGNet} 	    &0.8527	&0.6403	&0.6155	&35.1805	&\underline{6.4861}	&1.6979	&0.5810 & 0.8022    & 0.5698    & 0.4132    & 34.7651    & 6.8447    & 1.1492     & 0.4815    \\ \midrule
    ALBEF-CAM\cite{ALBEF}  &0.8405 	&0.5706 	&0.6135 	&34.4579 	&7.0563 	&1.7987 	&0.5374  &0.7990 	 &0.5783 	 &\textbf{0.5462} 	 &33.6213 	 &7.6827 	 &\underline{1.6559} 	 &0.5002  \\
    BLIP-CAM\cite{BLIP}    &0.8030 	&0.5736 	&0.5566 	&34.8112 &6.8114 	&1.7782 	&0.5014 	&0.7764 	&0.5765 	&0.4974 	&34.5738 &6.9772 	&1.6221 	&0.4685 \\ 
    ALBEF\cite{ALBEF}	    &0.8703	&0.6753	&0.6737	&35.1792	&6.5670	&1.7206	&\underline{0.6119} &0.8123	&0.6039	&0.4506	&34.4335	&7.0745	&1.2840	&0.4933 \\
    BLIP\cite{BLIP}	    &\underline{0.8765}	&\underline{0.6797}	&\underline{0.6784}	&35.1830	&6.5537	&1.7818	&0.6059 &\underline{0.8276}	&\underline{0.6147}	&0.4787	&34.0446	&7.3440	&1.4272	&0.4981 \\
    ImageBind\cite{ImageBind}	&0.8530	&0.6541	&0.6356	&\underline{35.2579}	&6.5018	&1.7700	&0.5786	 &0.7974	&0.5903	&0.4483	&34.7305	&6.8686	&1.2483	&0.4975 \\				
    TGSal	    &\textbf{0.8765}	&\textbf{0.6818}	&\textbf{0.7331}	&\textbf{35.2627}	&\textbf{6.4815}	&\textbf{1.8294}	&\textbf{0.6231} & \textbf{0.8371} & \textbf{0.6308} & \underline{0.5400}   & \textbf{34.9674} & \textbf{6.7004} & \textbf{1.6745} & \textbf{0.5431} \\
    \bottomrule[1.5pt]
    \end{tabular}
    }
\vspace{-1em}    
\end{table*}

\begin{table*}[!t]
    
    \centering
    \caption{Quantitative comparisons between our proposed TGSal model and benchmark methods under the conditions of common descriptions containing both salient and non-salient objects and averaged among all conditions. All models are first pretrained on SALICON \cite{SALICON_database}, and then jointly finetuned on all conditions of SJTU-TIS.}
    \label{SALICON_jointly_both_ave}
    \vspace{-0.8em}
    \setstretch{1.1}
    \resizebox{0.73\textwidth}{!}{
    \begin{tabular}{l|ccccccc|ccccccc}
    \toprule[1.5pt]
    Type    & \multicolumn{7}{c|}{Common descriptions contain both salient and non-salient objects }   & \multicolumn{7}{c}{Averaged among all conditions}     \\ \midrule
    Model\textbackslash{}Metirc & AUC-J↑ & sAUC↑  & CC↑    & IG↑     & KL↓     & NSS↑   & SIM↑   & AUC-J↑ & sAUC↑  & CC↑    & IG↑     & KL↓     & NSS↑   & SIM↑\\ \midrule
    SALICON\cite{SALICON}      &0.8183	&0.6393	&0.6094	&34.9234&	6.7364	&1.3853	&0.5627 &0.8250	&0.6273	&0.6090	&34.8628	&6.7099	&1.4365	&0.5628 \\
    ML-Net\cite{ML-Net}      & 0.8131  & 0.5927  & 0.5274  & 34.8449  & 6.7711  & 1.4570   & 0.5138  & 0.8024 & 0.5812  & 0.5051  & 34.8195  & 6.7778  & 1.3927 & 0.5051  \\
    SalGAN\cite{SalGAN}      & 0.8530   & 0.6555  & 0.6404  & 35.1226  & 6.5786  & 1.6233  &0.6063 & 0.8575 & 0.6635  & 0.7022  & \underline{35.2070}   & 6.5009  & 1.7065 & 0.6367  \\
    SAM-VGG\cite{SAM}    & 0.8259  & 0.6116  & 0.6227  & 34.7354  & 6.8470   & 1.4914  & 0.6056  & 0.8421 & 0.6269  & 0.6802  & 34.9246  & 6.7049  & 1.6448 & 0.6380   \\
    SAM-ResNet\cite{SAM} & 0.8491  & 0.6164  & 0.6426  & 34.9624  & 6.6894  & 1.6946  & 0.6029  & 0.8565 & 0.6214  & 0.7023  & 35.0734  & 6.6017  & \underline{1.7741} & 0.6495  \\
    GazeGAN\cite{GazeGAN}     & 0.7971  & 0.6237  & 0.5408  & 34.7606  & 7.0283  & 1.1652  & 0.5677  & 0.8159 & 0.6443  & 0.5919  & 34.9198  & 6.9645  & 1.2936 & 0.5873  \\
    VQSal\cite{duan2022saliency}       & 0.8568  & 0.6455  & \underline{0.6541} & 34.9100    & 6.7260   & 1.7270  & 0.6051  & \underline{0.8593} & 0.6475  & \underline{0.7157}  & 34.9126  & 6.7132  & 1.7660  & \underline{0.6562}  \\
    TranSalNet\cite{TranSalNet}  & \underline{0.8600}   & 0.6484  & 0.6370   & \underline{35.1997} & \underline{6.5268} & 1.7195  & 0.6015  & 0.8555 & 0.6554  & 0.6559  & 35.1452  & 6.5451  & 1.7035 & 0.5975  \\
    TempSAL\cite{TempSAL}     & 0.8120   & 0.5984  & 0.5708  & 34.9045  & 6.7298  & 1.2584  & 0.5830   & 0.8308 & 0.6289  & 0.6420   & 35.0792  & 6.5977  & 1.4465 & 0.6150   \\
    GSGNet\cite{GSGNet}      & 0.8363  & 0.6171  & 0.6306  & 35.1097  & 6.5876  & 1.4994  & 0.6062  & 0.8489 & 0.6340   & 0.6868  & 35.1635  & \underline{6.4816}  & 1.6375 & 0.6425  \\ \midrule
    ALBEF-CAM\cite{ALBEF}                   &0.8102 	&0.5507 	&0.5335 	&34.2644 	&7.1735 	&1.6297 	&0.5136 &0.8134     &0.5620 &0.5542 	&34.2155 	&7.2176 	&1.6269 	&0.5133      \\
    BLIP-CAM\cite{BLIP}	 &0.7965 	&0.5595 	&0.5424 	&34.7055 	&6.8677 	&1.6078 	&0.5310   &0.7886 	&0.5637 	&0.5233 	&34.6879 	&6.8789 	&1.5687 	&0.4999 \\
    ALBEF\cite{ALBEF}  &0.8542	&0.6638	&0.6480	&34.9905	&6.6702	&1.6768	& \underline{0.6065} &0.8558	&0.6629	&0.7022	&34.9712	&6.6744	&1.6783	&0.6439\\
    BLIP\cite{BLIP} &0.8593	&\underline{0.6669}	&0.6435	&34.6702	&6.8922	&\underline{1.7874}	&0.6061	&0.8579	&\underline{0.6651}	&0.6958	&34.5209	&6.9847	&1.7370	&0.6321 \\
    ImageBind\cite{ImageBind} &0.8387	&0.6312	&0.6423	&35.1161	&6.5832	&1.5888	&0.6002 &0.8475	&0.6437	&0.7004	&35.1884
	&6.5082	&1.6834	&0.6388 \\
    TGSal       & \textbf{0.8731} & \textbf{0.6676} & \textbf{0.6678} & \textbf{35.3714} & \textbf{6.4231} & \textbf{1.9424} & \textbf{0.6185} & \textbf{0.8680} & \textbf{0.6724} & \textbf{0.7532} & \textbf{35.2993} & \textbf{6.4138} & \textbf{1.8770} & \textbf{0.6715} \\
    \bottomrule[1.5pt]
    \end{tabular}
    }
    \vspace{-1em}
\end{table*}

\begin{table*}[!t]
\vspace{-0.5em}
    
    \centering
    \caption{Quantitative comparisons between our proposed TGSal model and benchmark methods under the pure image condition and the general scenario description condition. All models are jointly trained on all conditions of SJTU-TIS without pretraining on SALICON \cite{SALICON_database}.}
    \label{NOSALICON_jointly_pure_whole}
    \vspace{-0.8em}
    \setstretch{1.1}
    \resizebox{0.73\textwidth}{!}{
    \begin{tabular}{l|ccccccc|ccccccc}
    \toprule[1.5pt]
    Type                        & \multicolumn{7}{c|}{Pure images}                                       & \multicolumn{7}{c}{General scenario descriptions}                                       \\ \midrule
    Model\textbackslash{}Metirc & AUC-J↑ & sAUC↑  & CC↑    & IG↑     & KL↓     & NSS↑   & SIM↑   & AUC-J↑ & sAUC↑  & CC↑    & IG↑     & KL↓     & NSS↑   & SIM↑\\ \midrule
    SALICON\cite{SALICON}     & 0.8165 & 0.6456 & 0.7084 & 34.5467 & 6.7807 & 1.4682 & 0.6084 & 0.8423 & 0.5612 & 0.6437 & 35.1675 & 6.4235 & 1.4358 & 0.5421 \\
    ML-Net\cite{ML-Net}      & 0.7866 & 0.5710  & 0.5423 & 34.7310  & 6.8192 & 1.3174 & 0.5560  & 0.8169 & 0.5724 & 0.5572 & 34.9965 & 6.6267 & 1.5185 & 0.5152 \\
    SalGAN\cite{SalGAN}      & 0.8358 & 0.6655 & 0.7316 & 34.9031 & 6.6999 & 1.4314 & 0.6694 & 0.8699 & 0.6823 & 0.7224 & 35.2543 & 6.4480  & 1.6232 & 0.6277 \\
    SAM-VGG\cite{SAM}    & 0.8471 & 0.6528 & 0.7725 & 34.9361 & 6.6770  & 1.6652 & \underline{0.7094} & 0.8795 & 0.6550  & 0.7961 & \underline{35.4609} & 6.4048 & 1.9117 & 0.6054 \\
    SAM-ResNet\cite{SAM} & 0.8516 & 0.6275 & 0.7750  & 34.6470  & 6.8774 & \underline{1.7542} & 0.7016 & 0.8854 & 0.6280  & 0.7945 & 35.4399 & \underline{6.3193} & 1.9013 & 0.6312 \\
    GazeGAN\cite{GazeGAN}     & 0.8372 & 0.6612 & 0.7402 & 34.8437 & 6.7411 & 1.4600   & 0.6828 & 0.8717 & 0.6875 & 0.7496 & 35.3834 & 6.3585 & 1.6848 & 0.6066 \\
    VQSal\cite{duan2022saliency}       & \underline{0.8623} & 0.6483 & \underline{0.7934} & 34.5678 & 6.8612 & 1.7532 & 0.7012 & \underline{0.8854} & 0.6663 & \underline{0.8015} & 35.1169 & 6.5941 & \underline{1.9423} & 0.6432 \\
    TranSalNet\cite{TranSalNet}  & 0.8493 & 0.6654 & 0.7285 & 35.0877 & 6.6503 & 1.5338 & 0.6556 & 0.8745 & 0.6803 & 0.7073 & 35.2873 & 6.5252 & 1.6816 & 0.6024 \\
    TempSAL\cite{TempSAL}     & 0.8272 & 0.6424 & 0.6653 & 34.8994 & 6.7025 & 1.2900   & 0.6230  & 0.8681 & 0.6650  & 0.6754 & 35.1479 & 6.5217 & 1.5110  & 0.5801 \\
    GSGNet\cite{GSGNet}      & 0.8433 & 0.6622 & 0.7675 & \underline{35.1628} & \underline{6.5199} & 1.5404 & 0.6964 & 0.8777 & 0.6762 & 0.7707 & 35.4586 & 6.4064 & 1.7794 & 0.6354 \\ \midrule
    ALBEF\cite{ALBEF} &0.8361	&0.6647	&0.7641	&32.9646	&8.0436	&1.5410	&0.6654 &0.8718	&0.6840	&0.7722	&34.1203	&7.2340	&1.7979	&0.6434 \\
    BLIP\cite{BLIP} &0.8265	&0.6650	&0.7649	&31.5062	&9.0545	&1.5889	&0.6582 &0.8761	&\underline{0.6876}	&0.7775	&34.1488	&7.2143	&1.7836	&0.6471\\
    ImageBind\cite{ImageBind} &0.8380	&\underline{0.6690}	&0.7546	&34.9790	&6.6473	&1.4998	&0.6819 &0.8751	&0.6830	&0.7724	&35.3983	&6.3482	&1.7588	&\underline{0.6475} \\
    TGSal       & \textbf{0.8853} & \textbf{0.6701} & \textbf{0.8052} & \textbf{35.3140}  & \textbf{6.4069} & \textbf{1.9543} & \textbf{0.7158} & \textbf{0.8875} & \textbf{0.6917} & \textbf{0.8147} & \textbf{35.4859} & \textbf{6.3082} & \textbf{1.9674} & \textbf{0.6560}  \\
    \bottomrule[1.5pt]
    \end{tabular}
    }
    \vspace{-1em}
\end{table*}

\begin{table*}[!t]
    
    \centering
    \caption{Quantitative comparisons between our proposed TGSal model and benchmark methods under the conditions of specified descriptions of salient objects and specified descriptions of non-salient objects. All models are jointly trained on all conditions of SJTU-TIS without pretraining on SALICON \cite{SALICON_database}.}
    \label{NOSALICON_jointly_salicon_nonsalicon}
    \vspace{-0.8em}
    \setstretch{1.1}
    \resizebox{0.73\textwidth}{!}{
    \begin{tabular}{l|ccccccc|ccccccc}
    \toprule[1.5pt]
    Type                        & \multicolumn{7}{c|}{Specified descriptions of salient objects}                                    & \multicolumn{7}{c}{Specified descriptions of non-salient objects}                                 \\ \midrule
    Model\textbackslash{}Metirc & AUC-J↑ & sAUC↑  & CC↑    & IG↑     & KL↓     & NSS↑   & SIM↑   & AUC-J↑ & sAUC↑  & CC↑    & IG↑     & KL↓     & NSS↑   & SIM↑\\ \midrule
    SALICON\cite{SALICON} & 0.8213 & 0.6320  & 0.5054 & 34.8534 & 6.9531 & 1.4357 & 0.4829 & 0.7623 & 0.5721 & 0.3923 & 34.2345 & 7.1542 & 1.1258 & 0.4452 \\
    ML-Net\cite{ML-Net} & 0.8036 & 0.5764 & 0.4410  & 34.8008 & 6.8186 & 1.3610  & 0.4537 & 0.7732 & 0.5598 & 0.3756 & 34.0049 & 7.0448 & 1.1634 & 0.4436 \\
    SalGAN\cite{SalGAN} & 0.8316 & 0.6542 & 0.5583 & 34.8083 & 6.8134 & 1.4105 & 0.5392 & 0.7807 & \underline{0.5782} & 0.4099 & 34.1281 & 7.2862 & 1.0820  & 0.4595 \\
    SAM-VGG\cite{SAM}    & 0.8235 & 0.6387 & 0.5613 & 34.6546 & 6.9199 & 1.5112 & 0.5404 & 0.7791 & 0.5700   & 0.3904 & 33.9018 & 7.4430  & 1.0597 & \underline{0.4639} \\
    SAM-ResNet\cite{SAM} & \underline{0.8566} & 0.6333 & 0.6221 & 34.7105 & 6.8812 & 1.7713 & \underline{0.5833} & \underline{0.7918} & 0.5712 & 0.4056 & 34.0833 & 7.3172 & 1.1763 & 0.4634 \\
    GazeGAN\cite{GazeGAN}& 0.8194 & 0.6462 & 0.5227 & 34.4274 & 7.0774 & 1.3348 & 0.5270  & 0.7594 & 0.5733 & 0.3543 & 33.8322 & 7.4913 & 0.9149 & 0.4537 \\
    VQSal\cite{duan2022saliency}  & 0.8432 & 0.6458 & 0.6264 & 34.9812 & 6.7531 & \underline{1.7823} & 0.5814 & 0.7863 & 0.5511 & 0.4024 & 34.0576 & 7.3875 & 1.1687 & 0.4544 \\
    TranSalNet\cite{TranSalNet}  & 0.8536 & 0.6511 & 0.5752 & \underline{35.0809} & 6.7127 & 1.5522 & 0.5243 & 0.7915 & 0.5671 & 0.4113 & 34.3462 & 6.9916 & 1.1987 & 0.4569 \\
    TempSAL\cite{TempSAL}& 0.8068 & 0.6271 & 0.4830  & 34.7747 & 6.8367 & 1.1726 & 0.4851 & 0.7448 & 0.5637 & 0.3420  & 34.3796 & 7.0311 & 0.8518 & 0.4456 \\
    GSGNet\cite{GSGNet} & 0.8278 & 0.6423 & 0.5525 & 35.0140  & \underline{6.6708} & 1.4187 & 0.5408 & 0.7720  & 0.5653 & 0.3683 & 34.3397 & \underline{6.9593} & 0.9686 & 0.4637 \\ \midrule
    ALBEF\cite{ALBEF} &0.8462	&\underline{0.6562}	&0.6048	&33.1970	&7.9303	&1.6454	&0.5568 &0.7771	&0.5738	&\underline{0.4177}	&30.3490	&9.9056	&1.2018	&0.4532 \\
    BLIP\cite{BLIP} &0.8491	&0.6557	&\underline{0.6291}	&32.9974	&8.0687	&1.6813	&0.5711	&0.7732	&0.5658	&0.4062	&30.0693	&10.099	&\underline{1.2108}	&0.4501 \\
    ImageBind\cite{ImageBind}  &0.8404	&0.6520	&0.6070	&34.9981	&6.6819	&1.5857	&0.5591  &0.7830	&0.5695	&0.4083	&\underline{34.5652}	&6.9832	&1.1318	&0.4573 \\
    TGSal  & \textbf{0.8591}	& \textbf{0.6615}	& \textbf{0.7243}	& \textbf{35.2446}	& \textbf{6.4941}	& \textbf{1.7983}	& \textbf{0.6414} & \textbf{0.8351}	& \textbf{0.6167}	& \textbf{0.5331}	& \textbf{35.0413}	& \textbf{6.6532}	& \textbf{1.6314}	& \textbf{0.5231} \\
    \bottomrule[1.5pt]
    \end{tabular}
    }
    \vspace{-1em}
\end{table*}

\begin{table*}[t]
    
    \centering
    \caption{Quantitative comparisons between our proposed TGSal model and benchmark methods under the conditions of common descriptions containing both salient and non-salient objects and averaged among all conditions. All models are jointly trained on all conditions of SJTU-TIS without pretraining on SALICON \cite{SALICON_database}.}
    \label{NOSALICON_jointly_both_ave}
    \vspace{-0.8em}
    \setstretch{1.1}
    \resizebox{0.73\textwidth}{!}{
    \begin{tabular}{l|ccccccc|ccccccc}
    \toprule[1.5pt]
    Type    & \multicolumn{7}{c|}{Common descriptions contain both salient and non-salient objects }   & \multicolumn{7}{c}{Averaged among all conditions}     \\ \midrule
    Model\textbackslash{}Metirc & AUC-J↑ & sAUC↑  & CC↑    & IG↑     & KL↓     & NSS↑   & SIM↑   & AUC-J↑ & sAUC↑  & CC↑    & IG↑     & KL↓     & NSS↑   & SIM↑\\ \midrule
    SALICON\cite{SALICON}     & 0.8045    & 0.6184    & 0.6072    & 34.7756    & 6.8234    & 1.2675    & 0.5438    & 0.8106    & 0.6125    & 0.5942    & 34.6874    & 6.8193    & 1.3669    & 0.5385    \\
    ML-Net\cite{ML-Net}      & 0.8102    & 0.5662    & 0.5194    & 34.8428    & 6.7726    & 1.4335    & 0.5140     & 0.7962    & 0.5695    & 0.4963    & 34.6845    & 6.8169    & 1.3519    & 0.5064    \\
    SalGAN\cite{SalGAN}      & 0.8139    & \underline{0.6235} & 0.5920     & 34.4906    & 7.0167    & 1.2899    & 0.5526    & 0.8280     & \underline{0.6449} & 0.6243    & 34.7479    & 6.8274    & 1.3781    & 0.5863    \\
    SAM-VGG\cite{SAM}    & 0.8129    & 0.5946    & 0.5901    & 34.3134    & 7.1395    & 1.3648    & 0.5746 & 0.8315    & 0.6273    & 0.6472    & 34.7005 & 6.8769    & 1.5296    & 0.6005 \\
    SAM-ResNet\cite{SAM} & 0.8399 & 0.5946    & 0.6312    & 34.3193    & 7.1354    & 1.5672 & \underline{0.5756} & \underline{0.8462} & 0.6137    & 0.6672    & 34.6412    & 6.9013 & 1.6541    & \underline{0.6095}    \\
    GazeGAN\cite{GazeGAN}     & 0.8006    & 0.6046    & 0.5486    & 34.2901    & 7.1557    & 1.1959    & 0.5678    & 0.8209    & 0.6390  & 0.6093    & 34.6034    & 6.9275    & 1.3417    & 0.5868    \\
    VQSal\cite{duan2022saliency}       & 0.8368 & 0.6155    & 0.6341 & 34.8511    & 6.8650     & \underline{1.5828} & 0.5751    & 0.8461 & 0.6292    & \underline{0.6752} & 34.6904    & 6.8870     & \underline{1.6638} & 0.6094 \\
    TranSalNet\cite{TranSalNet}  & \underline{0.8405} & 0.6033    & 0.6200      & 34.8602 & \underline{6.7082} & 1.5308    & 0.5709    & 0.8431    & 0.6388    & 0.6285 & 34.9583    & 6.7064    & 1.5052 & 0.5776    \\
    TempSAL\cite{TempSAL}     & 0.7887    & 0.5903    & 0.5123    & 34.7064    & 6.8671    & 1.0851    & 0.5341    & 0.8105    & 0.6218    & 0.5572    & 34.8012 & 6.7769    & 1.2001    & 0.5485    \\
    GSGNet\cite{GSGNet}      & 0.8045    & 0.5814    & 0.5450     & 34.8422 & 6.7730  & 1.2083    & 0.5667    & 0.8281    & 0.6316    & 0.6286    & \underline{34.9967}    & \underline{6.6416} & 1.4093    & 0.5999    \\ \midrule
    ALBEF\cite{ALBEF} &0.8254	&0.6128	&\underline{0.6438}	&32.4258	&8.4479	&1.5067	&0.5627 &0.8321	&0.6427	&0.6611	&32.6702	&8.2675	&1.5390	&0.5912\\
    BLIP\cite{BLIP} &0.8209	&0.6186	&0.6374	&32.1613	&8.6312	&1.4639	&0.5674	&0.8287	&0.6430	&0.6633	&32.0649	&8.6871	&1.5529	&0.5920 \\
    ImageBind\cite{ImageBind} &0.8243	&0.6115	&0.6421	&\underline{34.9024}	&6.7312	&1.4389	&0.5611 &0.8331	&0.6423	&0.6565	&34.9703	&6.6732	&1.4858	&0.5981 \\
    TGSal  & \textbf{0.8740}	& \textbf{0.6660}	& \textbf{0.7001}	& \textbf{35.3922}	& \textbf{6.4087}	& \textbf{2.0492}	& \textbf{0.5899} & \textbf{0.8711}	& \textbf{0.6627}	& \textbf{0.7304}	& \textbf{35.2987}	& \textbf{6.4463}	& \textbf{1.8925}	& \textbf{0.6403} \\
    \bottomrule[1.5pt]
    \end{tabular}
    }
    \vspace{-1em}
\end{table*}

\vspace{-1em}
\subsection{Comparison with State-of-the-art Methods on Our Text-guided Image Saliency Database}
\label{model_train2}
We further conduct experiments on our SJTU-TIS database to validate the effectiveness and superiority of our proposed model on the text-guided image saliency prediction tasks. We compare the proposed TGSal model with ten classical saliency models including IT \cite{IT}, AIM \cite{AIM}, GBVS \cite{GBVS}, SMVJ \cite{SMVJ}, SUN \cite{SUN}, Hou \cite{Hou_NIPS}, SeR \cite{SeR}, CA \cite{CA}, HFT \cite{HFT}, CovSal \cite{CovSal}, ten DNN saliency models including SALICON \cite{SALICON}, ML-Net \cite{ML-Net}, SalGAN \cite{SalGAN}, SAM-VGG \cite{SAM}, SAM-ResNet \cite{SAM}, GazeGAN \cite{GazeGAN}, VQSal \cite{duan2022saliency}, TranSalNet \cite{TranSalNet}, TempSAL \cite{TempSAL}, GSGNet \cite{GSGNet} and three text-image pretraining models including ALBEF \cite{ALBEF}, BLIP \cite{BLIP} and ImageBind \cite{ImageBind}. The text-image pretraining models are added from two perspectives for comparison. Firstly, inspired by the class activation map \cite{CAM}, we directly use the cross-attention map (CAM) in text-image pretraining models for text-related image saliency prediction, specifically looking at how it highlights text-related image regions. We test two CAM-based methods on two text-image pretraining models including BLIP \cite{BLIP} and ALBEF \cite{ALBEF}, which are represented using BLIP-CAM \cite{BLIP} and ALBEF-CAM \cite{ALBEF}. Secondly, we use three text-image pretraining models as feature extraction methods, and then connect them to our decoder for training.

\subsubsection{Pretrained on SALICON and Finetuned Individually on Different Conditions of SJTU-TIS}
It should be noted that all of the DNN models are first pretrained on SALICON \cite{SALICON_database} and then finetuned on the five groups of our SJTU-TIS database. Table \ref{pure and whole} shows the quantitative comparisons of different models under the pure image condition and the general scenario description condition. Table \ref{salient and non-salient} presents the quantitative comparisons of different models under the condition of specified descriptions of salient objects and the condition of specified descriptions of non-salient objects. Table \ref{common} demonstrates the quantitative comparisons of different models under the condition of common descriptions containing both salient and non-salient objects and averaged among all conditions. In these tables, TGSal-I means leaving the text blank and only inputting images to the network, while TGSal means the complete network. Detailed analyses are given below.

Firstly, as shown in Table \ref{pure and whole}, in the case of pure images and general scenario descriptions, the unimodal baseline models, especially SAM-VGG also attain good performance. However, as shown in Tables \ref{salient and non-salient} and \ref{common}, these models have relatively poor performance, especially in terms of the CC and SIM metrics. These phenomena indicate that the text-guided saliency prediction is a difficult task for the unimodal baseline models. Moreover, the performance of all models under the conditions of the salient-object description and the common description containing both salient and non-salient objects is relatively better than under the non-salient-object description condition. This is mainly due to that the salient objects generally occupy a larger or more central area in the image, while non-salient objects are usually small objects. Generally, it is difficult for deep neural networks to extract features for small objects compared to large objects, which makes it difficult to concentrate on non-salient object.
We can also observe that deep learning models achieve better performance than traditional models in most cases, and mutimodal model performs better in most conditions among the baseline deep learning models.

Secondly, from the quantitative perspective, our proposed model achieves the best performance on all five subsets, verifying the great representation of learning ability under different description scenarios, and when the text is left blank, our model TGSal-I still outperforms other unimodal baseline models. The qualitative results shown in Fig. \ref{output} also manifest the superiority of our TGSal. Moreover, comparing the TGSal-I and TGSal model, we can observe that the TGSal model outperforms TGSal-I under all text-induced conditions. This phenomenon manifests that the introducing text information can improve the performance of predicting text-guided visual saliency, and our proposed text-image feature fusion modules can effectively introduce text information into the image saliency prediction task. To quantify the performance improvement, we compare TGSal (utilizing both image and text as input) with TGSal-I (only image input) under four conditions, respectively. As illustrated in Table \ref{grow}, the performance improvement percentage on the non-salient description set (Type3) is significantly greater than that in other categories, indicating that the text features provide more assistance for the saliency prediction under the ``specific descriptions of non-salient objects'' condition. However, the influence of text features on Type1 (general scenario descriptions) and Type4 (common descriptions containing both salient and non-salient objects) is comparatively modest.

\subsubsection{Pretrained on SALICON and Jointly Finetuned on All Conditions of SJTU-TIS}
We further conduct experiments considering the joint training strategy. Specifically, we first pretrain the model on SALICON \cite{SALICON_database}, and then jointly finetune it on all conditions of SJTU-TIS, which can further eliminate the  description category learning bias and improve the generalization capability of the TGSal model. The results are shown in Tables \ref{SALICON_jointly_pure_whole}, \ref{SALICON_jointly_salicon_nonsalicon} and \ref{SALICON_jointly_both_ave}. Since the cross-attention map cannot be used in the pure image condition, we only report the results of ALBEF-CAM \cite{ALBEF} and BLIP-CAM \cite{BLIP} on Type1 to Type4 in Tables \ref{SALICON_jointly_pure_whole}, \ref{SALICON_jointly_salicon_nonsalicon}, and \ref{SALICON_jointly_both_ave}. First, we observe that the CAM-based saliency prediction method for the multimodal text-image pretraining models can generate significant results, but the performance of most metrics is not comparable to the state-of-the-art methods. The finetuned three multimodal-based methods including ALBEF \cite{ALBEF}, BLIP \cite{BLIP}, and ImageBind \cite{ImageBind} can achieve the comparable performance, and may even perform better in terms of some metrics, compared to unimodal saliency prediction methods. Moreover, as a multimodal saliency prediction method, our TGSal achieves better performance compared to other three multimodal models, \textit{i.e.,} ALBEF \cite{ALBEF}, BLIP \cite{BLIP}, and ImageBind \cite{ImageBind}. This is reasonable since BLIP \cite{BLIP} is a pretrained model designed for image and text visual question and answering (VQA), and ImageBind \cite{ImageBind} is a pretrained model that contains more modalities besides image and text modalities. Their features contain more redundant information compared to CLIP \cite{radford2021learning_34}, which makes it hard to train to select right features.
Finally, we observe that the overall performance of the jointly trained model is slightly lower than the performance of the individually trained models, shown in Tables \ref{pure and whole}, \ref{salient and non-salient}, and \ref{common}, which is mainly due to the description mixing.

\subsubsection{Jointly Trained on All Conditions of SJTU-TIS without Pretraining on SALICON}
In order to validate that the proposed database
can independently support training and evaluation without the help of other databases, we further train all models jointly on all conditions of the SJTU-TIS dataset without pretraining on SALICON dataset \cite{SALICON_database}, and show the results in Tables \ref{NOSALICON_jointly_pure_whole}, \ref{NOSALICON_jointly_salicon_nonsalicon} and \ref{NOSALICON_jointly_both_ave}. It can be observed that when training directly on the SJTU-TIS dataset without pretraining on SALICON \cite{SALICON_database}, our proposed model achieves significant results and still performs better than other models in terms of all metrics, which validates that the proposed database can independently support training and evaluation. However, the performance of all models has decreased to some extend compared to the results after pretraining on SALICON \cite{SALICON_database} in Tables \ref{SALICON_jointly_pure_whole}, \ref{SALICON_jointly_salicon_nonsalicon} and \ref{SALICON_jointly_both_ave}.

\vspace{-1em}
\subsection{Ablation Studies}
\subsubsection{Contribution of the Text Feature Fusion Module}
To verify the rationality and effectiveness of the text feature fusion modules in the proposed TGSal model, we conduct ablation experiments from the following two aspects, including \emph{w/o} global which means removing the global text concatenation part of the global text feature fusion (GTFF) module, and \emph{w/o} local which means removing the attention module in the local text feature fusion (LTFF) module. Table \ref{Ablation} demonstrates the results of this ablation study, which manifests that these two modules are indispensable. Moreover, the results of \emph{w/o} local indicate that local text fusion is more crucial compared to the global text fusion in our proposed framework. Moreover, the comparison between TGSal-I and TGSal in Tables \ref{pure and whole}, \ref{salient and non-salient}, and \ref{common} also demonstrates the significance of introducing text features in the text-guided saliency prediction task.

\begin{table}[t]
    \centering
    \caption{The ablation study of the text feature fusion module.}
    \vspace{-0.8em}
    \setstretch{1.15}
    \resizebox{0.43\textwidth}{!}{
    \begin{tabular}{c|ccccccc}
    \toprule[1.5pt]
    Model\textbackslash{}Metric & AUC-J↑ & sAUC↑  & CC↑    & IG↑     & KL↓    & NSS↑   & SIM↑   \\ \midrule
    \emph{w/o} global              & 0.8525  & 0.6293 & 0.5635 & 35.0238 & 6.5237 & 1.7632 & 0.5502 \\
    \emph{w/o} local              & 0.8405  & 0.6393 & 0.5485 & 35.0019 & 6.6805 & 1.6703 & 0.5448 \\
    Proposed                       & \textbf{0.8601}  & \textbf{0.6405} & \textbf{0.6210}  & \textbf{35.3746} & \textbf{6.4222} & \textbf{1.9288} & \textbf{0.5774} \\ \bottomrule[1.5pt]
    \end{tabular}
    }
    \label{Ablation}
    \vspace{-1em}
\end{table}

\begin{figure*}[!t]
\centering
\vspace{-1em}
\includegraphics[width=6in]{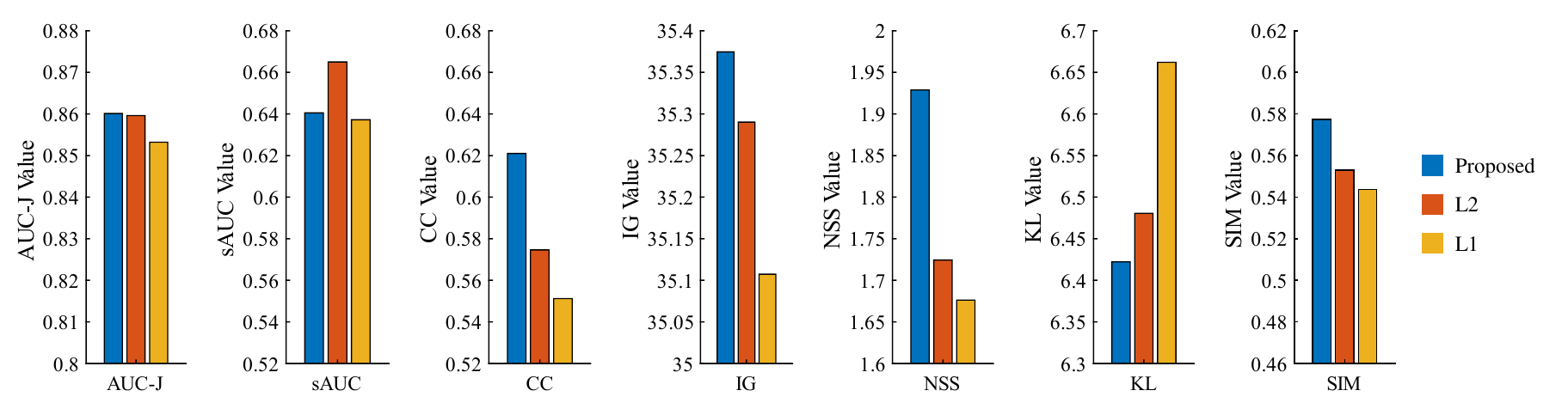}
\vspace{-1em}
\caption{The influence of different loss functions on the saliency prediction performance.}
\vspace{-1em}
\label{loss}
\end{figure*}

\begin{figure}[!t]
\centering
\includegraphics[width=3in]{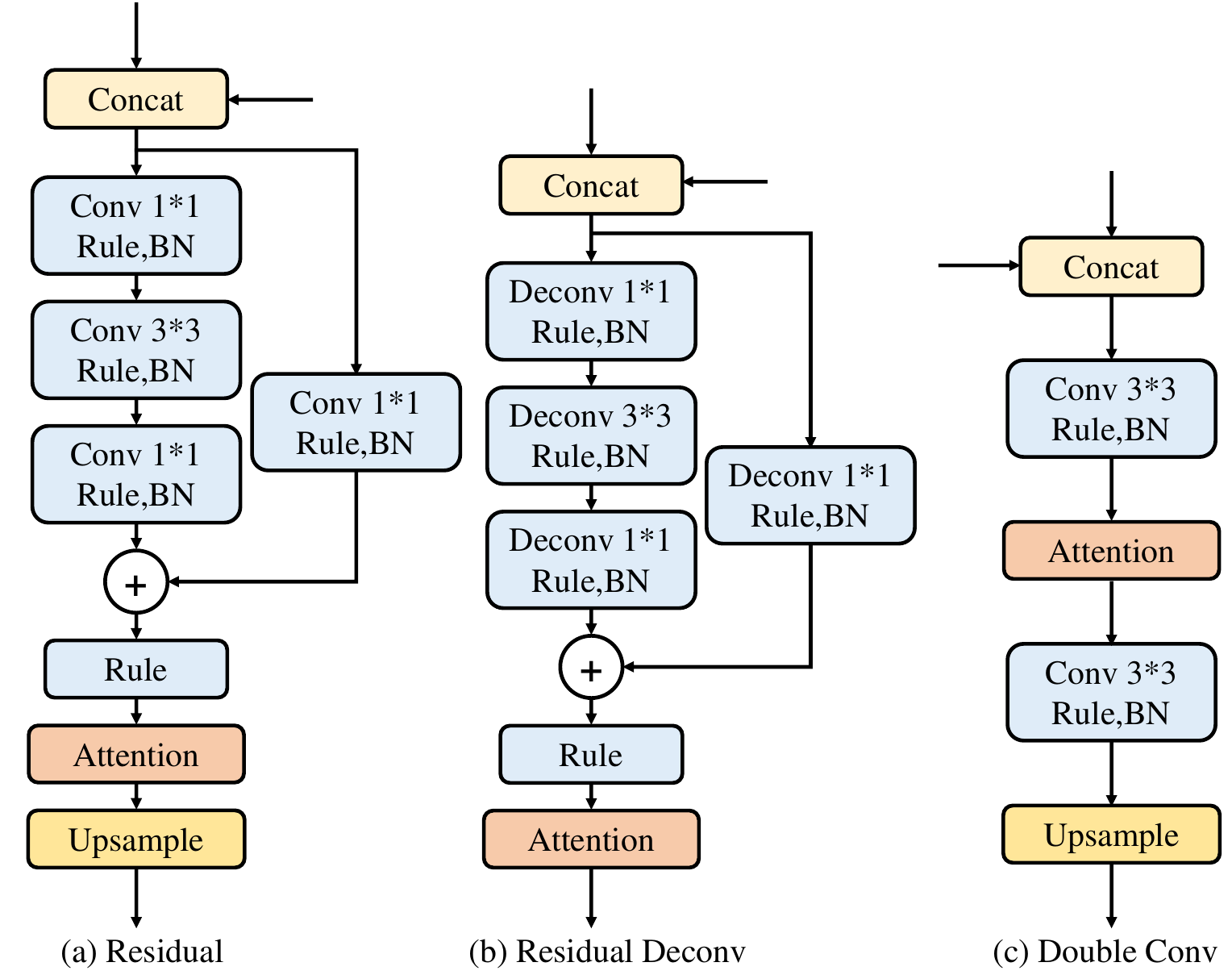}
\vspace{-1em}
\caption{Different feature fusion structures. (a) Residual. (b) Residual Deconv. (c) Double Conv.}
\label{mulImage}
\vspace{-1em}
\end{figure}

\begin{table}[t]
    \centering
    \caption{The influence of the number of heads in the attention module. (``Baseline'' represents the best performance results of the state-of-the-art baseline models from the TABLE \ref{salicon}.)}
    \vspace{-1em}
    \setstretch{1.15}
    \resizebox{0.43\textwidth}{!}{
    \begin{tabular}{c|ccccccc}
    \toprule[1.5pt]
    Model         & AUC-J↑ & sAUC↑  & CC↑    & IG↑     & KL↓    & NSS↑   & SIM↑   \\ 
    \midrule
    Baseline & \underline{0.8627}  & \underline{0.6578} & 0.8688 & \underline{35.2235} & \underline{5.4170} & 1.8599 & 0.7653  \\
    Number 2       & 0.8578  & 0.6473 & 0.8757 & 35.1892 & 5.6392 & 1.8604 & 0.7662 \\
    Number 4         & 0.8585  & 0.6365 & \underline{0.8801} & 34.3850 & 5.9905 & \textbf{1.8841} & \underline{0.7700} \\
    Number 8         & \textbf{0.8658}  & \textbf{0.6650} & \textbf{0.8816}  & \textbf{35.2529} & \textbf{5.3767} & \underline{1.8606} & \textbf{0.7731} \\ 
    \bottomrule[1.5pt]
    \end{tabular}
    }
    \label{head}
    \vspace{-1em}
\end{table}

\begin{table}[t]
    \centering
    \caption{The exploration of different image feature fusion structure. (``Baseline'' represents the best performance results of the state-of-the-art baseline models from the TABLE \ref{salicon}.)}
    \vspace{-0.8em}
    \setstretch{1.15}
    \resizebox{0.43\textwidth}{!}{
    \begin{tabular}{c|ccccccc}
    \toprule[1.5pt]
    Model                 & AUC-J↑ & sAUC↑  & CC↑    & IG↑     & KL↓    & NSS↑   & SIM↑   \\ \midrule
    Baseline         & 0.8627  & 0.6578 & 0.8688 & \underline{35.2235} & 5.4170 & 1.8599 & \underline{0.7653}  \\
    Residual            & 0.8642  & 0.6578 & 0.8683 & 34.8502 & 5.4623 & 1.8593 & 0.6293 \\
    Residual Deconv  & 0.8329  & 0.6387 & 0.6900   & 34.8272 & 5.8273 & 1.8273 & 0.6082\\
    Double Conv     & \underline{0.8650}   & \underline{0.6623} & \underline{0.8728} & 34.9745 & \underline{5.3934} & \textbf{1.8728} & 0.6476            \\ 
    TGSal                 & \textbf{0.8658}  & \textbf{0.6650}  & \textbf{0.8816} & \textbf{35.2529} & \textbf{5.3767} & \underline{1.8606} & \textbf{0.7731} \\ \bottomrule[1.5pt]
    \end{tabular}
    }
    \label{residual}
    \vspace{-1em}
\end{table}

\begin{table}[!t]

    \centering
    \caption{Comparison between different dataset partitions. In each type, the first row means that the dataset is divided into a training set, a validation set, and a testing set respectively, to conduct experiments. The second row means that the dataset divided into a training set and a testing set respectively, to conduct experiments. The two testing sets are the same for each type.}
    \label{mydataset311}
    \vspace{-0.8em}
    \setstretch{1}
    \resizebox{0.33\textwidth}{!}{
    \begin{tabular}{c|ccccccc}
    \toprule[1.5pt]
    Type\textbackslash Metric   & AUC-J↑  & sAUC↑    & CC↑    & NSS↑     & SIM↑     \\ \midrule
          &  0.8666        &    0.6497      &   0.8674       &  1.9130         &  0.7649       \\
    \multirow{-2}{*}{Pure image} & {0.8672} & {0.6551} & {0.8736} & {1.9093} & {0.7711} \\ \midrule
            & 0.8921         &  0.7035        &   0.8370       &  2.1058         &   0.7299               \\
    \multirow{-2}{*}{Type 1}   & {0.8946} & {0.7057} & {0.8501} & {2.0916} & {0.7166} \\ \midrule
          &  0.8699        &  0.6786        &  0.6559        &  1.8914         &   0.5997             \\
    \multirow{-2}{*}{Type 2}    & {0.8728} & {0.679}  & {0.6702}  & {1.9615} & {0.6023} \\ \midrule
          &  0.8536        &    0.6404      &  0.6014        &  1.8385         &   0.5555           \\
    \multirow{-2}{*}{Type 3}     & {0.8601} & {0.6405} & {0.6210}  & {1.9288} & {0.5774} \\ \midrule
          &   0.8662       &  0.6641        &   0.7240       &   1.8373        &    0.6617            \\
    \multirow{-2}{*}{Type 4}   & {0.8684} & {0.6793} & {0.7317}  & {1.8449} & {0.6696} \\ 
    \bottomrule[1.5pt]
    \end{tabular}
    }
    \vspace{-1em}
\end{table}

\subsubsection{Comparison between Different Head Numbers of Attention}
Since the number of heads in the attention module is a hyperparameter, we also conduct an ablation experiment to study the impact of the number of heads on our TGSal model based on the SALICON database. As shown in Table \ref{head}, we compare the results of three different numbers of heads including 2, 4, and 8, respectively. The baseline represents the best performance result among all baseline deep learning models from Table \ref{salicon}. We bold the best results in this table. It can be observed that when the number of heads is 2 and 4, some metrics are lower than the baseline model, when the number of heads is set to 8, all metrics are higher than the baseline model and most of them achieve better performance compared to the third line. Since increasing the head number may also add the GPU memory consumption, the head number in our TGSal is set to 8.

\subsubsection{Comparison between Different Image Feature Fusion Structures}
Due to the use of a series of operations such as convolution and upsampling in the feature fusion module to fuse multi-scale image information, we also explore the impact of the design of the feature fusion module based on the SALICON database. We attempt three structures, as shown in Fig. \ref{mulImage}. Specifically, the first structure uses residual modules instead of the convolution module in the LTFF module and HFR module of our TGSal model. The second structure uses deconvolution structures instead of convolutional layers and upsampling layers. The third structure attempts to replace one convolution with two, and insert an attention module in the middle of the convolutions. The experimental results shown in Table \ref{residual} demonstrate that the deconvolution structure has the poorest performance, while the third structure achieves the best in terms of a few metrics. However, the performance of the above three structures in term of SIM is much lower than that of the baseline model. Therefore, the final structure of the LTFF module and HFR module are designed as shown in Fig. \ref{model}.

\subsubsection{Comparison between Different Loss Functions}

To further validate the effectiveness of the adopted loss function, we compare it with other common loss functions: L1 norm loss and mean square error loss. Fig. \ref{loss} shows the results of each metric under the condition of non-salient description. 
Despite observing that the proposed and used loss function exhibits relatively lower values in terms of the sAUC metric, the performance is higher than other two loss functions in terms of other metrics. Since it is important to consider the performance across a comprehensive set of metrics and our presented loss function achieves better performance compared to the traditional L1 and L2 loss functions in terms of most metrics, we adopt this new loss function in our work. The better performance of L2 loss in terms of the sAUC metric may come from the explicit prediction loss since sAUC is an accuracy metric, but it may not work well on other similarity-based metrics.

\subsubsection{Comparison between Different Database Partitions}
The above experiments are all conducted on the condition that the database is split into training/testing sets with the ratio of 4:1. We further add the experiment containing a validation set for comparison. 
We split the database into training/testing sets with the ratio of 4:1, and into training/validation/testing sets with the ratio of 3:1:1 for each type, respectively, and compare the results. Table \ref{mydataset311} demonstrates the comparison results between different database partitions.
It can be observed that for the condition that we split the data into three sets, there is a minor performance drop for most cases and metrics, but for a few cases and metrics, the performance may be improved. The performance difference is not large, which manifests the stability of our model. However, the slight performance drop for most cases may be due to the reduction of the training samples.

\begin{figure}[!t]
\centering
\includegraphics[width=3.4in]{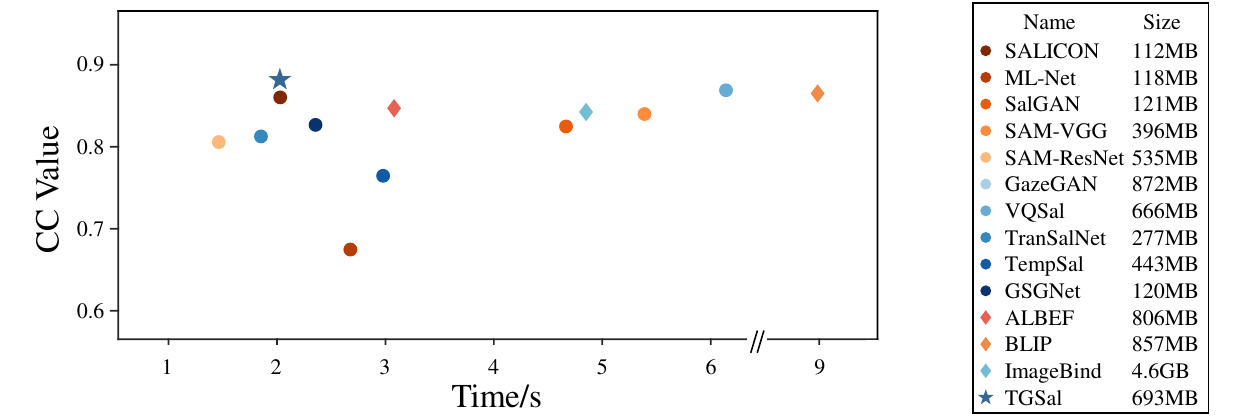}
\vspace{-1em}
\caption{Scatter plots of the weighted average CC results versus the running time for each saliency prediction method on the SALICON dataset.
The model size of each method is also reported on the right side.}
\vspace{-1em}
\label{run_speed}
\end{figure}

\subsection{Model Running Speed and Size Comparison}
Since computational efficiency and model size are crucial in practical applications, we also compare the computational efficiency and model size of different saliency prediction models. We test the running speed of all these models on a server with an 8255C CPU @ 2.50GHz and an NVIDIA GeForce RTX 2080 Ti graphic card, operating on Ubuntu. We report the model sizes and running times in Fig. \ref{run_speed}. The running time of each model is calculated on 100 images with the resolution of 640$\times$480, and is the average of 10 repeated tests. It can be observed that our proposed model achieves the best performance while exhibiting a highly efficient running speed compared to other models. Although some unimodal saliency models have relatively small sizes, their performance is not good. Moreover, the model sizes of the multimodal models are larger than our model, while the proposed TGSal achieves better performance.

\section{Conclusion}
\label{Conclusion}
Visual attention analysis and prediction are important tasks in multimedia systems. In this work, we conduct an in-depth exploration of text-induced visual attention and saliency prediction. Specifically, we construct the first text-guided image saliency database termed SJTU-TIS, where an image has multiple different text descriptions. Our constructed SJTU-TIS database contains 1200 text-image pairs and the correspondingly collected eye movement data. Through qualitative and quantitative analysis, we conclude that text descriptions do have influence on the visual attention, and different types of text descriptions of the same image may have different influences on the corresponding visual attention, which mainly depends on the objects being described. A novel text-guided saliency prediction model, termed TGSal, is then proposed to better predict the text-guided image saliency, which extracts both text and image features and hierarchically fuses them during the decoding process. Experimental results on the SALICON database and the SJTU-TIS database validate that our proposed method outperforms the benchmark saliency prediction models under both pure image and text-guided conditions, demonstrating the superiority and generality of the model. Moreover, under the conditions of the text descriptions of objects, the performance of our proposed TGSal is significantly improved with introducing the text features into the backbone compared to without using them, therefore manifests the importance of the text-image feature fusion for the text-guided saliency prediction task.

\bibliographystyle{IEEEtran}
\bibliography{main}

\begin{IEEEbiography}[{\includegraphics[width=1in,height=1.25in,clip,keepaspectratio]{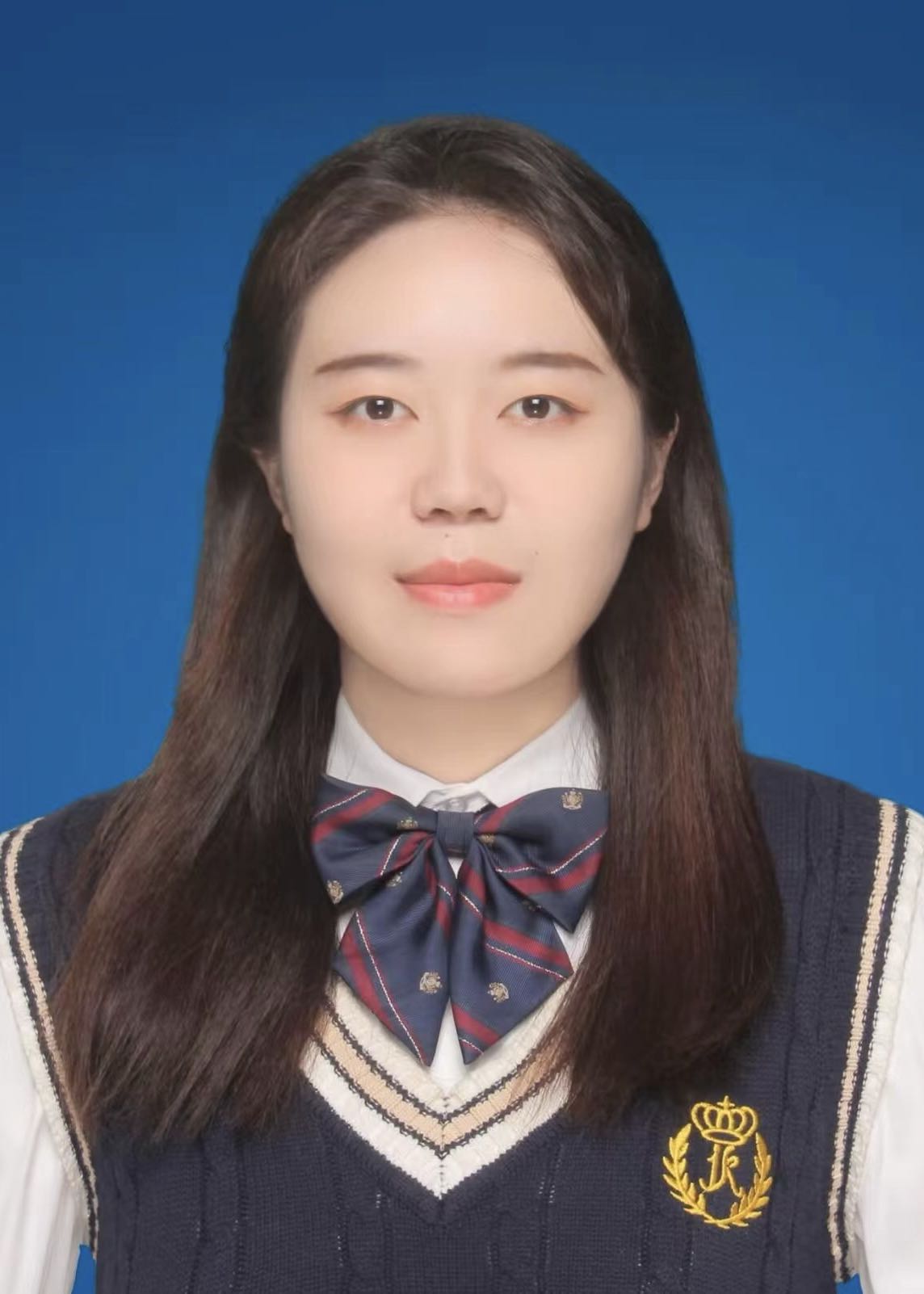}}]{Yinan Sun}
received the B.E. degree from Wuhan University, Wuhan, China, in 2022. She is currently pursuing her Ph.D. degree at the School of Electronic Information and Electrical Engineering at Shanghai Jiao Tong University, Shanghai, China. Her research interests include multimodal visual attention modeling and multimedia signal processing.
\end{IEEEbiography}

\begin{IEEEbiography}[{\includegraphics[width=1in,height=1.25in,clip,keepaspectratio]{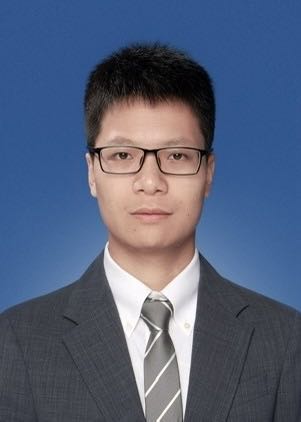}}]{Xiongkuo Min}
(Member, IEEE) received the B.E. degree from Wuhan University, Wuhan, China, in 2013, and the Ph.D. degree from Shanghai Jiao Tong University, Shanghai, China, in 2018, where he is currently a tenure-track Associate Professor with the Institute of Image Communication and Network Engineering. From Jan. 2016 to Jan. 2017, he was a visiting student at University of Waterloo. From Jun. 2018 to Sept. 2021, he was a Postdoc at Shanghai Jiao Tong University. From Jan. 2019 to Jan. 2021, he was a visiting Postdoc at The University of Texas at Austin and the University of Macau. He received the Best Paper Runner-up Award of IEEE Transactions on Multimedia in 2021, the Best Student Paper Award of IEEE International Conference on Multimedia and Expo (ICME) in 2016, the excellent Ph.D. thesis award from the Chinese Institute of Electronics (CIE) in 2020, and the Best Paper Award of IEEE International Symposium on Broadband Multimedia Systems and Broadcasting (BMSB) in 2022, and several first place awards of grand challenges held at IEEE ICME and ICIP. His research interests include image/video/audio quality assessment, quality of experience, visual attention modeling, extended reality, and multimodal signal processing.
\end{IEEEbiography}

\begin{IEEEbiography}[{\includegraphics[width=1in,height=1.25in,clip,keepaspectratio]{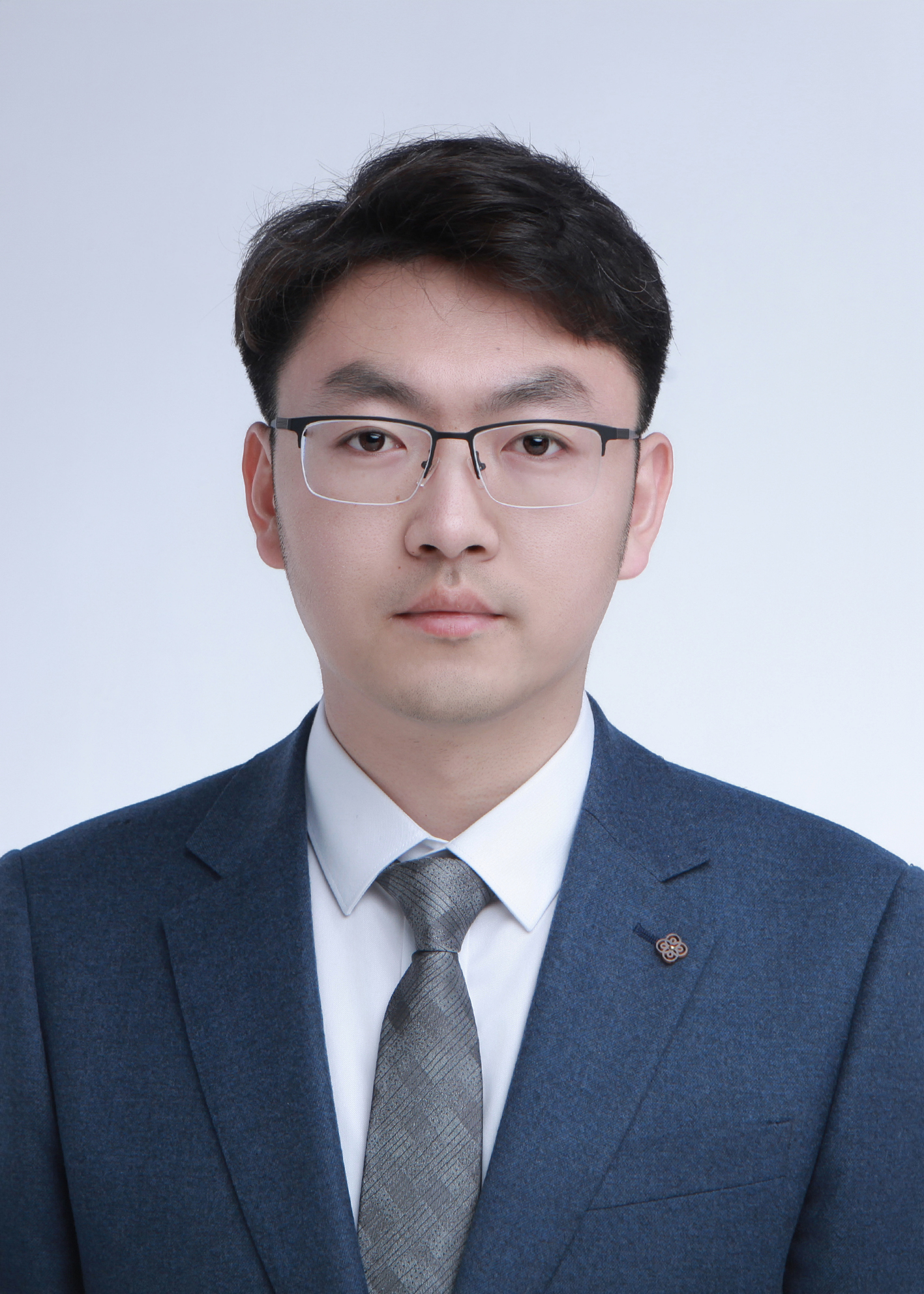}}]{Huiyu Duan}
received the B.E. degree from the University of Electronic Science and Technology of China, Chengdu, China, in 2017, and the Ph.D. degree from Shanghai Jiao Tong University, Shanghai, China, in 2024. He is currently a Postdoctoral Fellow at Shanghai Jiao Tong University. From Sept. 2019 to Sept. 2020, he was a visiting Ph.D. student at the Schepens Eye Research Institute, Harvard Medical School, Boston, USA. He received the Best Paper Award of IEEE International Symposium on Broadband Multimedia Systems and Broadcasting (BMSB) in 2022. His research interests include perceptual quality assessment, quality of experience, visual attention modeling, extended reality (XR), and multimedia signal processing.
\end{IEEEbiography}

\begin{IEEEbiography}[{\includegraphics[width=1in,height=1.25in,clip,keepaspectratio]{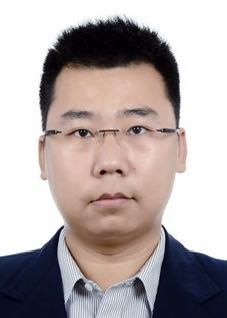}}]{Guangtao Zhai}
(Senior Member, IEEE) received the B.E. and M.E. degrees from Shandong University, Shandong, China, in 2001 and 2004, respectively, and the Ph.D. degree from Shanghai Jiao Tong University, Shanghai, China, in 2009. From 2008 to 2009, he was a Visiting Student with the Department of Electrical and Computer Engineering, McMaster University, Hamilton, ON, Canada, where he was a Postdoctoral Fellow from 2010 to 2012. From 2012 to 2013, he was a Humboldt Research Fellow with the Institute of Multimedia Communication and Signal Processing, Friedrich Alexander University of Erlangen-Nuremberg, Germany.
He is currently a Research Professor with the Institute of Image Communication and Information Processing, Shanghai Jiao Tong University. His research interests include multimedia signal processing and perceptual signal processing. He received the Award of National Excellent Ph.D. Thesis from the Ministry of Education of China in 2012.
\end{IEEEbiography}

\end{document}